\documentclass[11pt]{article}

\usepackage[final]{ACL2023}
\usepackage{times}
\usepackage{latexsym}
\usepackage{graphicx}
\usepackage{amsmath}
\usepackage{cuted}
\usepackage{mathtools, amssymb, lipsum}
\usepackage{cancel}
\usepackage[colorinlistoftodos]{todonotes}
\usepackage{enumitem}

\usepackage[symbol]{footmisc}

\usepackage[T1]{fontenc}
\usepackage[utf8]{inputenc}

\usepackage{microtype}

\usepackage{hyperref}
\usepackage{lineno}
\usepackage{comment}
\usepackage{amsmath}
\usepackage{float}
\usepackage{graphicx}
\usepackage{svg}
\usepackage{pdfpages} 
\modulolinenumbers[5]
\usepackage{pgfplots}
\usepackage{amssymb}
\usepackage{multirow}
\usepackage{multicol}
\usepackage{makecell}
\usepackage{caption}
\usepackage{subcaption}
\usepackage{graphics}

\usepackage{makecell}

\newcommand*\samethanks[1][\value{footnote}]{\footnotemark[#1]}
\def\@fnsymbol#1{\ensuremath{\ifcase#1\or \dagger\or \ddagger\or
   \mathsection\or \mathparagraph\or \|\or **\or \dagger\dagger
   \or \ddagger\ddagger \else\@ctrerr\fi}}
    \makeatother

\title{Imagination is All You Need! \\ Curved Contrastive Learning for Abstract Sequence Modeling \\ Utilized on Long Short-Term Dialogue Planning}
\author{Justus-Jonas Erker  \\
  DFKI \thanks{\; German Research Center for Artificial Intelligence}\; Lab Berlin \&  \\
  Maastricht University  \\ 
  \texttt{j.erker@student.} \thanks{\; maastrichtuniversity.nl} 
   \And
  Stefan Schaffer   \\
   DFKI \samethanks[1]\; Lab Berlin \\
    \texttt{stsc10@dfki.de} 
    \And
    Gerasimos Spanakis  \\
   Maastricht University  \\
   \texttt{jerry.spanakis@}\samethanks[2]
   } 

\begin{document}
\maketitle
%--> Paper should be split:
%\newline
%1. Paper: LTP Embeddings 
%\newline
%2. Paper: Safer Language with Semantic Search and Multi-Task Embeddings
%\newline
%3. Paper: ImagineGPT
%\newline
%--> first two would net cross-referencing though.
\begin{abstract}
%\todo{For Cambridge admission emphasize that two authors are supervisor --> they want "single-authored sample work" }
%Current State of the Art Dialogue Systems use the   
Inspired by the curvature of space-time \citep{Einstein1921-EINRTS-2}, we introduce \textbf{C}urved \textbf{C}ontrastive \textbf{L}earning (CCL), a novel representation learning technique for learning the \textbf{relative} turn distance between utterance pairs in multi-turn dialogues. The resulting bi-encoder models can guide transformers as a response ranking model towards a goal in a zero-shot fashion by projecting the goal utterance and the corresponding reply candidates into a latent space. Here the cosine similarity indicates the distance/reachability of a candidate utterance toward the corresponding goal. Furthermore, we explore how these forward-entailing language representations can be utilized for assessing the likelihood of sequences by the entailment strength i.e. through the cosine similarity of its individual members (encoded separately) as an emergent property in the curved space. These non-local properties allow us to imagine the likelihood of future patterns in dialogues, specifically by ordering/identifying future goal utterances that are multiple turns away, given a dialogue context. As part of our analysis, we investigate characteristics that make conversations (un)plannable and find strong evidence of planning capability over multiple turns (in 61.56\% over 3 turns) in conversations from the DailyDialog \citep{li-etal-2017-dailydialog} dataset. Finally, we show how we achieve higher efficiency in sequence modeling tasks compared to previous work thanks to our relativistic approach, where only the last utterance needs to be encoded and computed during inference.

\end{abstract}

\section{Introduction}\label{sec:intro}
\begin{figure}[ht]
    \centering
    \includegraphics[width=0.5\textwidth]{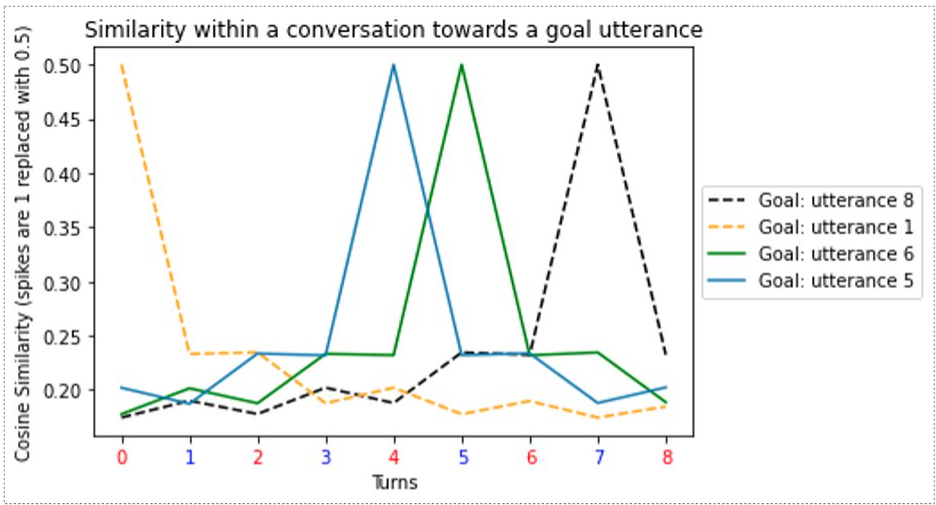}
    \caption{Entailment property of sentence transformer-based embeddings within conversations on DailyDialog}
    \label{fig:entail}
    % Important: explain blue and red color speaker
\end{figure}

Large Scale Transformers are becoming more and more popular in dialogue systems (\citet{DialogGPT}, \citet{peng2022godel}). Though these models are very effective in generating human-like responses in a given context, based on their learning objective to minimize perplexity, they tend to have trouble generating engaging dialogues \citep{DBLP:journals/corr/abs-2009-06978}. \citet{DBLP:journals/corr/abs-2202-00666} have shown that human conversations usually do not sample from the most likelihood of words like transformers do.  
We argue that one reason for this is that natural conversations can be (always) considered goal-oriented (even chitchat) and motivate this claim based on literature from psychology. These have shown that "Conversation is a goal-directed process" \citep{MYLLYNIEMI1986147} as humans shift conversation topics based on the social connection/audience and use it to shape social relations \citep{articleRelation}. 
%Language as an emergent property of human sociocultural evolution, 
%This social relationship component is crucial, even in chit-chat conversations with strangers, where the goal might only be to create a positive emotion in the corresponding individual.  
The psychological literature also elaborates on how humans are able to plan and simulate dialogues by utilizing inner speech as part of verbal working memory  \citep{grandchamp:hal-02290943}. 
\begin{quote}"Key to most of such models is that inner speech is posited as part of a speech production system involving predictive simulations or “forward models” of linguistic representations" \citep{AldersonDay2015InnerSD}
\end{quote} 
Keeping this in mind, we investigated dialogues under the aspect of "forward" entailing language representations by projecting them into a simple semantic sentence transformer \citep{DBLP:journals/corr/abs-1908-10084} latent space. We place a fixed position in the DailyDialog \citep{li-etal-2017-dailydialog} dataset as a goal utterance and measure the cosine similarity of the goal to every other utterance within the dialogue. Our own preliminary work revealed, as shown in figure \ref{fig:entail}, that the similarity of previous utterances to the goal utterance increases as they get closer to the goal utterance. 
 
 \begin{figure*}[ht]
    \centering
    \includegraphics[width=0.7\textwidth]{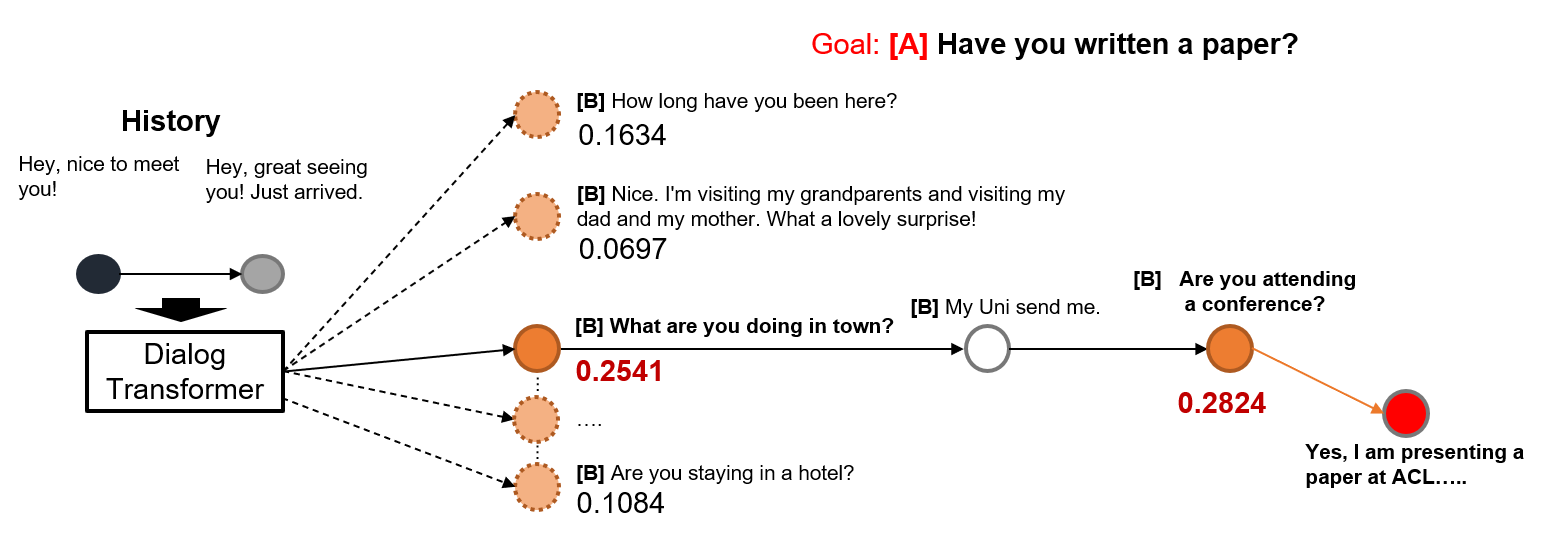}
    \caption{DialoGPT Guidance Example with Imaginary Embeddings with before \texttt{[B]} and after \texttt{[A]} token.}
    \label{fig:GuidanceExample}
\end{figure*}
However, fluctuations between the speaker at the goal turn (saying the utterance later on) and their dialogue partner can be observed. As we see on the blue \& red highlighted turns, the goal turn speaker has a greater similarity to the goal utterance than the dialogue partner. %\todo{Mir ist hier nicht ganz klar wer der Target Speaker und wer der Dialog Partner ist. Ist der Dialog Partner der Chatbot?}% 
We filtered all samples causing these fluctuations and find that these transitive entailing properties are essential for guiding the conversation toward the given goal. Regardless of whether the person had the intent to reach the target goal.%and furthermore give more control to the user who will generate the utterance later on. This is crucial as we do not want a dialogue system relying on the guidance of the dialogue partner. 
We demonstrate in this paper how we can build upon this phenomenon to learn the relative distance between utterance pairs. In particular, by mixing the training objective of Natural Language Inference (NLI) for the semantic embedding space with a distance proportional and directional aware (through two special tokens [BEFORE] \& [AFTER]) cosine similarity-based loss of utterance pairs. 
% add somewhere here forward entailing language representations; likelihood = strength of entailment 

The resulting Curved Contrastive Learning (CCL) is presented on three tasks: \textbf{(1)} short-term planning, \textbf{(2)} next utterance selection, and \textbf{(3)} long-term planning.

\textbf{(1) Short-term planning:} CCL allows us to imagine the likelihood of a candidate utterance leading to a given goal utterance by projecting them together into one latent space (imaginary space). The cosine similarity indicates the distance/reachability of a candidate utterance towards the corresponding goal as illustrated in a transformer guidance example in figure \ref{fig:GuidanceExample}. Thanks to the transitive property we can select the utterances at each turn greedily.
%Similar to the concept of transitivity of textual entailment graphs \citep{Kotlerman2015TextualEG} we can pick the next utterance towards our desired goal in a greedy way. 

\textbf{(2) Next utterance selection:} The embeddings can be utilized for sequence modeling by only using the cosine similarity between the separately encoded sequence members. It is evaluated by the ranking performance of the human vs random utterances task given a dialogue context.

\textbf{(3) Long-term planning:} Since these embeddings do not require entire sequences for sequence modeling, we can assess the likelihood of following patterns (of multiple goal utterances that are multiple turns apart) by using the entailment strength between these and the context in the curved space. We evaluate this approach based on the ordering/identifying of future goal utterances. 

Furthermore, we investigate two research questions:
\setlist{nolistsep}
\begin{itemize}[noitemsep]
\itemsep0em 
    \item  Do chit-chat conversations have planning capability? \textbf{(RQ1)}
    \item  What characteristics make dialogue planning possible? \textbf{(RQ2)}
\end{itemize}

The paper is structured as follows: In  §\ref{sec:relatable} we discuss the related work. Following in §\ref{sec:methods} where we present the methodology, baselines as well as basic components for the advanced architectures. In §\ref{sec:STP} the short-term planning approaches, followed by the next utterance selection in §\ref{sec:Imaginary Attenion} and the long-term planning approaches for ordering goals in §\ref{sec:ltp}. We  wrap up the paper with the experiments \& discussion in §\ref{sec:eval} followed by the conclusion in §\ref{sec:conclusion}. 

%As part of our analysis, we will provide a detailed analysis of characteristics when planning in dialogues is possible and where not.
%In the appendix, we will provide an outlook for our forward-entailing representations for other text-generation tasks, in particular story generation TODO. 

\section{Related Work}\label{sec:relatable}
Our work builds upon two major concepts, dialogue planning, and entailment. Related publications from the stated fields are discussed below. 

\subsubsection*{Dialogue Planning}
%Dialogue planning has a long tradition in task/goal-oriented settings and originally started with Dialogue Trees \citep{ChatbotsDialogueSystems} and over the years moved to more automatic Dialogue path generations \citep{articlePlanning}. %\citet{DBLP:journals/corr/abs-2009-12506} proposed that decoupling the process of planning and realization performs better than end-to-end approaches. 
While previously introduced planning techniques used several abstraction approaches \citep{articlePlanning}, none of them exploited the characteristics of curved conversation embedding latent spaces. We argue that generating a complete dialogue path is unnecessary as we can simply choose the utterance in the transformer's search space that gets us closest to the goal at every turn. \citet{https://doi.org/10.48550/arxiv.2205.07352} proposed a similar idea on word level by applying constrained decoding to the dialogue response generation to increase the likelihood of a target word not only in the current utterance but also utterances in the future. Furthermore, DialogRPT \citep{DBLP:journals/corr/abs-2009-06978} has been introduced as a dialogue response ranking model for depth, width, and upvotes prediction for utterance candidates. We utilize DialogRPT as a baseline for our next utterance selection experiments based on the dialogue history. %Todo ConVRT

\subsubsection*{Entailment}
Entailment-based approaches have a long history in NLP and have been utilized for a lot of tasks as zero-shot classification tasks like relation extraction \citep{obamuyide-vlachos-2018-zero} or zero-shot text classification \citep{DBLP:journals/corr/abs-1909-00161}. The idea of entailment graphs and making use of transitivity has been previously explored by 
\citet{Kotlerman2015TextualEG} \& \citep{https://doi.org/10.48550/arxiv.2204.03286}. Textual entailment has also been applied to Dialogue Systems as an evaluation technique \citep{EvaluatingCoherence} or for improving response quality through backward reasoning \citep{DBLP:journals/corr/abs-2105-00079}. Contrastive learning with positional information has been previously applied to image segmentation \citep{DBLP:journals/corr/abs-2106-09157}. While \citet{NEURIPS2020_3fe23034} utilized contrastive learning with augmentations for graph neural networks (GNNs). 
Natural Language Inference (NLI) based transformers have been increasingly used for semantic textual similarity (STS) since the introduction of Sentence Transformers, thanks to bi-encoders \citep{DBLP:journals/corr/abs-1908-10084} that can compare sentence pairs with cosine similarity and therefore reduce computation time by a $234000$ \footnote{According to \citet{DBLP:journals/corr/abs-1908-10084} a set of $10000$ Sentences would require 50 million inference computations with Bert which would, according to them, require 65 hours, while SBERT prior encoded would only take 5 seconds} fold. This trend has especially been supported by GPU Search \citep{DBLP:journals/corr/JohnsonDJ17}. These sentence transformers have successfully been applied to learn utterance representations for retrieving utterance replies in dialogue systems \citep{https://doi.org/10.48550/arxiv.2109.12599} or ConvRT \cite{henderson-etal-2020-convert} that we use as a baseline. However, without utilizing the curved property of conversations which we argue, as motivated in §\ref{sec:intro}, is essential for forward representations.% Last but not least Albert Einstein 

%We will demonstrate in this paper 

%With this paper, we contribute a new type of embedding that can be utilized for goal-oriented settings by mixing. Furthermore, encode its sequence member independently but  novel sequence modeling technique

%--> Say based on this nobody has done this: 
%+ add following paper reference: https://arxiv.org/abs/1804.07754 \citep{https://doi.org/10.48550/arxiv.1804.07754}

%+ following paper:  https://arxiv.org/abs/1911.03688 %\citep{https://doi.org/10.48550/arxiv.1911.03688} 

%\textbf{Key contributions: in Introduction!}
%\begin{itemize}
%    \item novel sequence modeling technique 
%    \item goal-oriented event prediction
%    \item planning in dialogues
%\end{itemize}

\section{Methods}\label{sec:methods}
In this section, we formally define the research questions (problem definition), our baselines for the evaluation, and the core of Imaginary Embeddings based on which advanced architectures are built in the following sections.
\subsection{Problem Definition Planning}
As part of this paper, we investigate two planning problems, short- and long-term planning. Short-term planning aims at guiding the conversation from the current position towards a given goal utterance $g$ (which we define as a semantic utterance) over multiple turns. Long-term planning, on the other hand, targets the ordering/scheduling of a set of goals $G$ (utterances that are multiple turns apart) within a conversation.

%\subsection{Baselines}
%The following baselines are considered:
% e directly

% Jerry: add paper
\subsection{Long-Short Term Planning Evaluation}\label{sec:LSTPE}
As part of this paper, we introduce a new evaluation technique, 
\textbf{L}ong-\textbf{S}hort \textbf{T}erm \textbf{P}lanning \textbf{E}valuation (\textbf{LSTPE}). LSTPE is split into Short- as well as Long-Term planning. 

\subsubsection{Short-Term Planing Evaluation}\label{sec:STPE}
As part of the short-term planning evaluation, we evaluate the guidance capability of imaginary embeddings towards a given goal utterance. For this purpose, we split all dialogues within a given corpus $d \in C$ into subsets of $d[:h_l]$ which represents the history of utterances (or context) with a fixed length $h_l$, $d[h_l]$ the "correct" following utterance and  $d[h_l+g_d]$ as goal utterance with a goal distance $g_d$. We then let a dialogue transformer generate 100 candidate utterances given the context $d[:h_l]$ for every dialogue $d \in C$ which we project together with the goal utterance into the imaginary embedding. Following, we compare the ranking score of the original utterance to the artificially generated utterances. As metrics, we report the Hits@$K$ ratio ($X$\%) and the average rank.

\subsubsection{Long-Term Planning Evaluation}\label{sec:LTPE}
Similar to the Short-Term planning,  we take a corpus of dialogue data $d \in C$ and split it at fixed positions $x$ into the dialogue history and three goal utterances $|G| = 3$. Given a dialogue history of length $h_l$,  $\forall d \in C: d[:h_l], d[x], d[x+g_d], d[x+2g_d]$ where $g_d\geq2$ is the distance between the goals. We define the first goal in distance as  $x - h_l$ in the perspective of the dialogue history. The three resulting goal utterances result in $6$ possible order permutations. Since 4 of them are partially ordered, we split the evaluation into ranking the partially ordered and reverse order to the true order separately. In both cases, we present the Hits@$K$ ratio ($X$\%) as well as the average total rank.
While this technique is simple and does not require any supervision, some samples due to the random selection are without any context indistinguishable. E.g. an utterance like "oh, okay" could be at any position. Since all models are evaluated on the same data set, this is not an issue, however, an accuracy of 100\% is realistically not possible. 

\subsection{Next Utterance Selection Evaluation}\label{sec:NextE}
Furthermore, we test the embedding's capability of telling potential replies from random utterances given a dialogue context by comparing it to DialogRPT \citep{DBLP:journals/corr/abs-2009-06978}, ConvRT \cite{henderson-etal-2020-convert} and BM25 \cite{10.1561/1500000019} on a ranking task. The data set is built up in a similar way as for short-term planning.

% rewrite that <

\subsection{Imaginary Embeddings with Curved Contrastive Learning}
\label{sec:IE}
\begin{figure}[ht]
    \centering
    \includegraphics[width=0.49\textwidth]{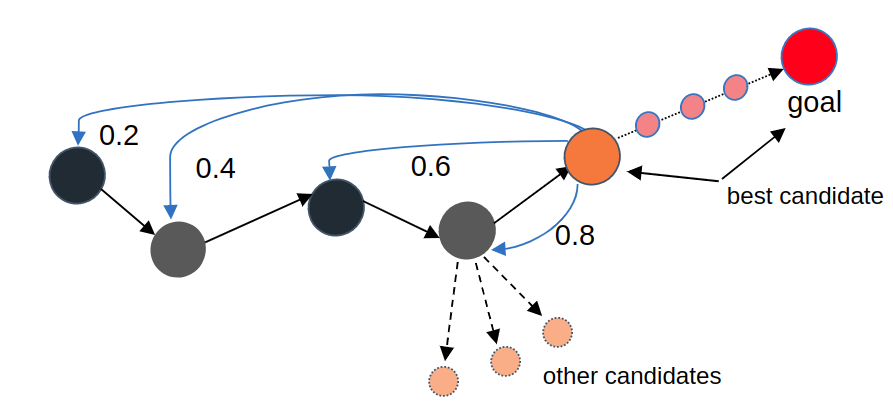}
    \caption{Curved property of Imaginary Embeddings. Grey/black nodes represent history utterances, orange nodes are utterance candidates, and dark orange is the best candidate as it is closest to the goal utterance (red). From the perspective of the best candidate encoded as \texttt{[A]}, the scores towards history illustrate the training objective as they are encoded with \texttt{[B]} tokens.}
    % explain red steps 
    % candidates emphasize more
    \label{fig:IEpath}
\end{figure}

% Original:
%We introduce a novel self-supervised learning technique to generate semantically meaningful embeddings which have a conversation positional as well as directional awareness between utterance pairs. To generate these properties, we train a bi-encoder sentence transformer on two training objectives. The first objective builds upon the AllNLI dataset (a combination of SNLI \citep{https://doi.org/10.48550/arxiv.1508.05326} and MultiNLI \citep{https://doi.org/10.48550/arxiv.1704.05426}) with a simple Softmax Loss. To learn the graph structure of conversations, two special tokens \texttt{[BEFORE]} and \texttt{[AFTER]} are introduced. The model is (pre-)trained with a Cosine Similarity loss on DailyDialog \citep{li-etal-2017-dailydialog}, by sliding through conversational data with a fixed length $l=5$. Notably, we combine consecutive utterances of the same speaker. Based on this fixed length, the training data is constructed for a given window as follows:
We introduce a novel self-supervised learning technique to map sequences to a conversational space. To generate these properties, we train a bi-encoder sentence transformer on two training objectives. The first objective builds upon the AllNLI dataset (a combination of SNLI \citep{https://doi.org/10.48550/arxiv.1508.05326} and MultiNLI \citep{https://doi.org/10.48550/arxiv.1704.05426}) with a simple Softmax Loss. To learn the conversational space, two special tokens \texttt{[BEFORE]} and \texttt{[AFTER]} are introduced. The model is (pre-)trained with a Cosine Similarity loss on DailyDialog \citep{li-etal-2017-dailydialog}, by sliding through conversational data with a fixed length $l=5$. Notably, we combine consecutive utterances of the same speaker. Based on this fixed length, the training data is constructed for a given window as follows:

\begin{equation}
\label{equation:trainingdataconstruction}
    \forall i \in \{1,..,l\}: \begin{cases}
  ([B] \ u[0], [A] \ u[i], s= \frac{l-i}{l} 
\\
  ([B] \ u[i], [A] \ u[0], \  s= 0
  \\
  ([B] \ u[0], [A] \ u'[r], s= 0
   \\
  ([B] \ u'[r], [A] \ u[0], s= 0
\end{cases}
\end{equation}

where $[A]$ = \texttt{[AFTER]}, $[B]$ = \texttt{[BEFORE]}, $u$ the utterances in the observed window, $u'$ a set of random utterances, and $s$ the cosine similarity score. 
As \ref{equation:trainingdataconstruction} shows, the target cosine similarity for a positive sample pair is proportional to their positional distance in the dialogue (see illustration in figure \ref{fig:IEpath}). 
This lets us learn semantic properties between \texttt{[B]} \& \texttt{[B]} and \texttt{[A]} \& \texttt{[A]} as well as the curvature as a \textbf{relative time dimension} between utterance pairs in the space between \texttt{[B]} \& \texttt{[A]} representations.
%with which learn the \textbf{relative} turn distance between utterances pairs.
%The emergence of non-local properties between utterance, we will furthermore add that thanks to the relative time dimension between utterance pairs (with the before and after token) and their resulting non-locality, we can map sequences to a conversational surface.

Three hard negatives are introduced, the first ensures the directional property by swapping the \texttt{[BEFORE]} and \texttt{[AFTER]} token. The following two are selected from a special dataset of random utterances. %We find that ensuring the random utterances are distinct from utterances in the given window by a simple semantic cosine similarity improves performance. 
Figure \ref{fig:IEpath} unveils the widespread utility of imaginary embeddings. As shown, we can simply pick the best candidate utterance for reaching a given goal by \textbf{imagining} the closeness of the candidate utterance to the goal in the curved space without requiring the \textbf{real} representations between the utterance pairs.

Similar to an object in our universe that always moves on a straight line but is curved by space-time \citep{Einstein1921-EINRTS-2}, we can follow a line to our goal utterance by greedily selecting the best utterance on turn-to-turn bases. We illustrated this transitive property by the light red in-between nodes in figure \ref{fig:IEpath}.

% original:
% While in the short planning the candidate utterances are sampled from a dialog transformer, we can simply ignore the closeness of candidate utterances to the history. In long-term planning, however, we can exploit the curved property of context utterances for goal ordering as the next goal should be the closest to the context utterances. Analogous applies to the best next utterance in the next utterance selection/ranking task.

Thanks to the relative time dimension between utterance pairs and their resulting \textbf{non-locality}, we are able to encode all sequence members (utterances) independently into one latent space and accumulate the likelihood of a sequence by comparing only with cosine similarity. In particular, by imagining the closeness between every context utterance (encoded with [B]) and the future utterance (encoded with [A])), i.e. Imagination is all you need!

Not only can we assess the likelihood of sequences that we explore in the next utterance selection §\ref{sec:Imaginary Attenion} but we can also utilize these self-organizing properties for mapping sequential representations to the conversational surface that are multiple turns apart. We explore this as the ordering of goals in long-term planning §\ref{sec:ltp}.

\subsubsection{Adding Speaker Tokens}
Furthermore, we can modify imaginary embeddings with additional speaker tokens. Given a multi-turn dialogue with two participants, the tokens \texttt{[O]} and \texttt{[E]} are added to the \texttt{[BEFORE]} utterance at the encoding step (for even and odd distances to the target utterance \texttt{[AFTER]}). Accordingly, the learning objective (see equation \ref{equation:trainingdataconstruction}) for the curved property is slightly modified by adding hard negatives for false speaker matches (see appendix \ref{sec:STLO}). 

\section{Short Term Planning Approach (Transformer Guidance)}\label{sec:STP}
As described in section \ref{sec:STPE} we utilize imaginary embeddings as a re-ranking model. Respectively, we let a task-specific dialogue transformer generate 100 candidate utterances given the context $d[:h_l]$ of a fixed length $h_l$ for every sample dialogue $d \in C$. To get a diverse distribution of utterances we choose nucleus sampling with $p=0.8$ and a temperature of $t=0.8$. The generated utterances from the transformer are then projected in the imaginary embedding space and the goal similarity of $d[h_l+g_d]$ is measured. Following, we check the rank of the true utterance from the test set leading to the goal utterance. The average rank and the distribution of ranks within the dialogue are evaluated with respect to different history lengths $h_l$ and different goal distances $g_d$. 

\section{Next Utterance Selection with Curving}\label{sec:Imaginary Attenion}
Motivated by the curved property, the most suitable next utterance $u_f \in U_F$ for a dialogue sequence $his$ should be closest to the individual utterances of the sequence on average. We can assess a relative likelihood between all future utterances by measuring the entailment strength $P_E$ (i.e. imagining the closeness) of every $u_f$ to the history of utterances based on the cosine similarity as follows:
\begin{equation}
    P_E(u_f| his) = \sum_{u_{i} \in his } \frac{{\bf [B] \; u_i}  \; \; {\bf [A] \; u_f}} {\|{\bf [B] \; u_i}\| \; \|{\bf [A] \; u_{f}}\|} 
\end{equation}
In the ranking evaluation, we sort the results of $\forall u_f \in U_F: P_E(u_f| his)$ to determine the rank of the true utterance.
Notably, we can observe the entailment strength (or activation) of individual utterances to a future one, which enables many other applications. During inference, while the dialogue partner is still speaking, we can pre-compute the entire context (apart from the new incoming utterance). Furthermore, we can utilize the curved context for greedily selecting the next goal  $\max\limits_{g \in G}P_E(g|his)$ in our long-term planning experiments. We refer to this as greedy curving.

%Applying this curving technique allows us to select the next goal greedily by simply encoding \textbf{only} every new incoming utterance, adding it to the dialogue history $his$, and selecting the next goal by

\section{Long-Term Planning Approaches} \label{sec:ltp}
In this section, we describe how Imaginary Embeddings can be used to order goals (a set of utterances) within dialogues for long-term planning. The models are evaluated with \textbf{LSTPE}, a given set of goals $G$ with $|G| = 3$, and an equal distance between each node. 
%\subsubsection*{Semantic Search}
%Accordingly, the semantic search index is constructed by concatenating $3$ embedding nodes with the corresponding distance from each other, creating  $3*384$ dimensional vectors: 
%\begin{equation}
%    \Vec{v} = [e[x], e[x+i], e[x+2i]]
%\end{equation}
%where $\Vec{v}$ is the index, furthermore, we define $O$ as the set of all possible orders $o$ over sequences of $e$. For every order $o$ we sample results from a joined corpus of DailyDialog, MovieCorpus as well as the Personachat Corpus. The following features are generated for every query: similarities with each of the three nodes between the query and the retrieved candidates, as well as the entailment property of the dialogue partner history or the entire dialogue history (like in Multi-Shot Attention, but without Imaginary Embeddings). Based on these features, a training set with 50\% positives (correct order) and 50\% negatives is constructed, on which a random forest is trained. 
%\subsubsection*{ImagineGPT Classifier}
%The ImagineGPT Classifier is trained on GPT NEO with a similar input as ImagineGPT, except that \texttt{[HEURISTICS]} are blanked out. The model gets a history within the  context token and then in equal distance utterances as goals of the same dialogue separated by the heuristic token. For negatives, we randomly order these goals. Like the Semantic Search, an evenly distributed training data set is generated. 

\subsection{Imaginary Embedding Chains}
 \begin{figure}[ht]
    \centering
    \includegraphics[width=0.49\textwidth]{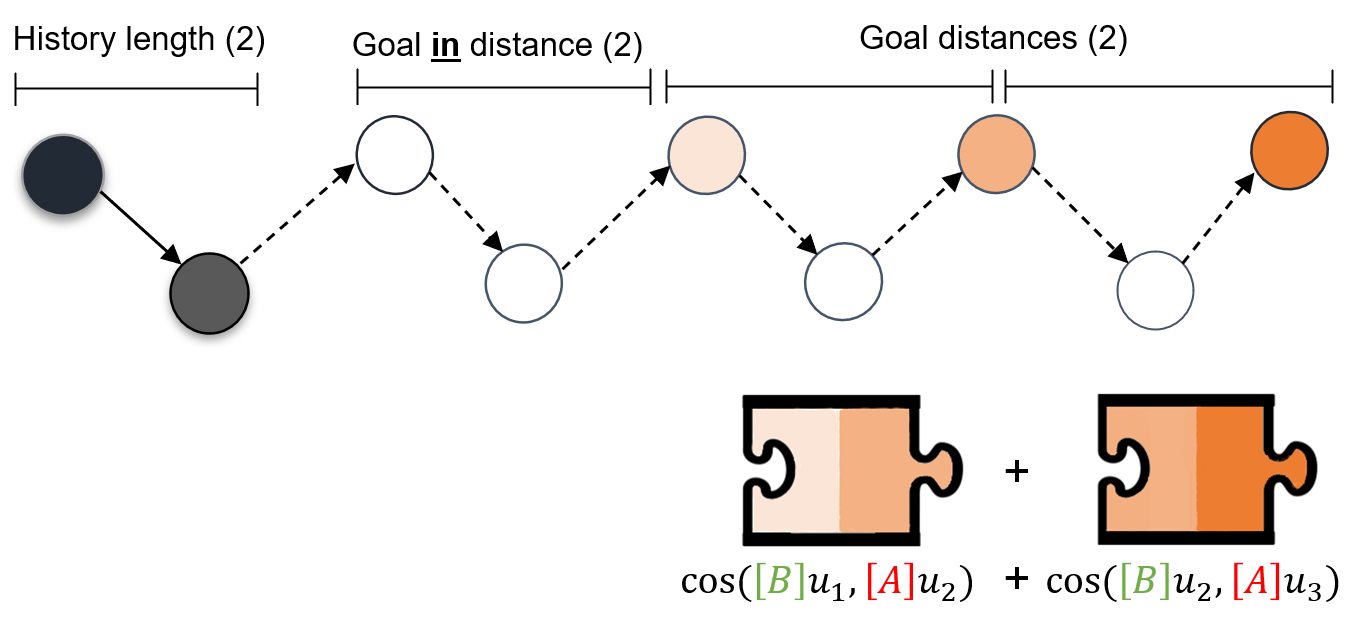}
    \caption{Long Term planning Dataset construction variables (history length, goal distances, (first) goal in distance) demonstrated. Furthermore, the concept of Imaginary Embedding Chains (IEC) is illustrated with its puzzle-like properties with the corresponding goal utterance colors.}
    \label{fig:dataset}
\end{figure}
Imaginary Embeddings are perfectly suited for this task as they can be concatenated into cosine similarity chains by using the (\texttt{[B]} before and \texttt{[A]} after token) as illustrated in figure \ref{fig:dataset}. We mathematically define it as:  
\begin{equation}\label{eq:chain1}
     s(o)= \Big(\sum_{i \in o} \frac{{{\bf [B] \; g_i}  \; \; {\bf [A] \; g_{i+1}}}}{\|{\bf [B] \; g_i}\| \; \|{\bf [A] \; g_{i+1}}\|} \Big)
\end{equation}
where we choose the order of goals $o \in O$ by the highest similarity score $s$ with $\underset{o \in O}{\max} (s(o))$ (strongest entailment strength) of a given sequence $o = <g_1,...,g_n>$ of goals $g_i \in G$. While this chain can be arbitrarily long and, thanks to GPU tensor computations calculated rather quickly, the complexity with $\mathcal{O}(n!)$ for a brute force computation remains high. 

%--> IMPORTANT CHANGE:
%entailment strength of history nodes to the future utterance assess the likelihood of that sequence. Other than a DialogCSE not fixed to a specific length.

\subsection{Imaginary Embedding Chains with History Curving}
Finally, we combine the concepts of Imaginary Embedding Chains and Curving by generating for every order $[g_1, g_2, g_3]$ a score (equation \ref{eq:hischain}):

\begin{multline}\label{eq:hischain}
s'(g_1, g_2, g_3)= s(o) + P_E(g_1|his) \\
    - \frac{1}{2}P_E(g_2|his)   - P_E(g_3|his) 
\end{multline}
where $s(o)$ is the chain score of the given order based on equation \ref{eq:chain1} and $ P_E(g_i|his)$ is the history curving score for the corresponding goal. We motivate the addition of $g_1$ and the subtraction of $g_3$ (as well as $g_2$) based on the presumption that $g_1$ should be closest while $g_3$ should be the furthest away to the history with respect to the curved property. Note that other than the simple Imaginary Embedding Chains (IEC), IEC + curving requires some dialogue context and is therefore not suitable for dialogue planning without context.

\section{Experiments}\label{sec:eval}
Our experiments are conducted on two dialogue corpora, DailyDialog \citep{li-etal-2017-dailydialog} and the Microsoft Dialogue Challenge (MDC) corpus \citep{https://doi.org/10.48550/arxiv.1807.11125}. We experiment with two transformer architectures BERT \citep{https://doi.org/10.48550/arxiv.1810.04805} and RoBERTa \citep{https://doi.org/10.48550/arxiv.1907.11692} to generate Imaginary Embeddings. In the short-term planning (transformer guidance) setting, we let our Imaginary Embeddings guide DialoGPT \citep{DialogGPT} for DailyDialog and GODEL \citep{peng2022godel} for the MDC corpus. For the next utterance selection, we use pre-trained checkpoints of DialogRPT \citep{DBLP:journals/corr/abs-2009-06978} and ConvRT \cite{henderson-etal-2020-convert} as baselines. Furthermore, we add BM25 \cite{10.1561/1500000019} as well as an ablation study with the two special tokens (before and after) but without the curved learning objective that we explore in the appendix \ref{sec:ab}. 

\subsection{Experimental Setup}
While the DailyDialog data set has a test corpus of 1000 dialogues, we first have to generate a test data set for MDC. We do so by extracting the last 333 samples for each of the three task-oriented domains (movie-ticket booking, restaurant reservation, and taxi booking). This leaves us with 11,118 dialogues as training data for DailyDialog and 9088 training samples for MDC. 

\subsection{Self-Supervised Training}
Apart from combining consecutive utterances of the same speaker and removing dialogues with utterances longer than 200 tokens, we apply no further pre-processing on the training data. As described in §\ref{sec:IE}, we pre-train all our architectures in stage (1) with a mixed training objective of NLI and the Curved Contrastive Learning (CCL) on the DailyDialog corpus for ~5 epochs. For all MDC models, we follow up with a second stage where we train on the target corpora with the curved property learning objective only for domain adaptation.% We find that the UserToken model for long-term planning on DailyDialog is the only DailyDialog model that improves with a second stage as well while all MDC models improve with DailyDialog pre-training. The number of ideal training steps in the last stage depends on the task.  
While Long Term planning performs best after ~5 epochs of further fine-tuning,  short-term planning requires only between 0.5 to 1 epoch(s). We provide all models including model cards on Huggingface as well as our code as part of a python package \texttt{pip install imaginaryNLP} (open-sourced under Apache-2.0 license) in the following GitHub repository \footnote{\url{https://github.com/Justus-Jonas/imaginaryNLP}}. 

\subsection{Evaluation Data sets}
The evaluation data sets DailyDialog and MDC are constructed analogously. We construct the datasets for STP based on history length and goal in distance \& LTP based on history length,  goal \textbf{in} distance, goal distances respectively as illustrated in figure \ref{fig:dataset}. Since MDC with an average number of 6.51 turns is even shorter than DialyDialog with 7.84, we are limited in the long-term planning to a shorter context as well as a goal in distance length. 

\begin{table*}[ht]
\centering

 %\medium{
    \renewcommand{\arraystretch}{1.2}
    
    \centering
    %\label{tab:STPEvalShort}
    \resizebox{\textwidth}{!}{
    
    \begin{tabular}{l | c c c c c | c c c c c}
    \Xhline{3\arrayrulewidth}
     & & & \multicolumn{6}{c}{}{\textbf{\thead{Human Utterance Ranking vs 100 utterances sampled \\ from DialoGPT Large / GODEL Large (p=0.8, t=0.8)}}} \\

      & & \multicolumn{3}{c}{}{\textbf{\thead{Imaginary Embedding \\ without Speaker Token}}} & & & \multicolumn{3}{c}{}{\textbf{\thead{Imaginary Embedding \\ with Speaker Token}}}  \\
    %\Xhline{3\arrayrulewidth}
    \multirow{2}{*}{\textbf{Goal in Distance}}& 
    \multirow{2}{*}{\textbf{\thead{Hits@5 \\ (in \%)}}} & 
    \multirow{2}{*}{\textbf{\thead{Hits@10 \\ (in \%)}}} & 
    \multirow{2}{*}{\textbf{\thead{Hits@25 \\ (in \%)}}} & 
    \multirow{2}{*}{\textbf{\thead{Hits@50 \\ (in \%)}}} &
    \multirow{2}{*}{\textbf{\thead{Average \\ Rank}}} &
    \multirow{2}{*}{\textbf{\thead{Hits@5 \\ (in \%)}}} & 
    \multirow{2}{*}{\textbf{\thead{Hits@10 \\ (in \%)}}} & 
    \multirow{2}{*}{\textbf{\thead{Hits@25 \\ (in \%)}}} & 
    \multirow{2}{*}{\textbf{\thead{Hits@50 \\ (in \%)}}} &
    \multirow{2}{*}{\textbf{\thead{Average \\ Rank}}} \\
    & & & & & & & & & & \\
    \Xhline{3\arrayrulewidth}
    %\Xhline{3\arrayrulewidth}
    %\multicolumn{8}{l}{\textbf{DailyDialogue Test Corpus even goal distance $g_d$ (saying goal by yourself)}} \\
     \textbf{DailyDialog Test Corpus} & & & & &  \\
         %& & & & & & & & & &  \\
        Guidance \textbf{even} $g$ distance  &      29.36 &       35.76 &       51.03 &        67.9 &             34.59 &          27.78 &           36.22 &           53.78 &           71.36 &                 32.56 \\

    %& 10 & 4 &  17.65 &   19.61 &   35.29 &   47.06 &  50.41 \\
    %& & & & & & & \\
    %\Xhline{3\arrayrulewidth}
     %& & & & & & & & & & \\
    
     Guidance \textbf{odd} $g$ distance &      31.31 &       39.21 &       54.09 &       72.78 &             30.61  &          \textbf{63.49} &           72.18 &           83.21 &           91.06 &                  \textbf{12.9} \\

    %& & & & & & & \\

    \Xhline{3\arrayrulewidth}

    \textbf{MDC Test Corpus} & & & & & & & \\
         %& & & & & & & & & & & & \\
        Guidance  \textbf{even} $g$ distance  &      20.79 &       29.32 &       48.04 &       70.85 &             34.86 &            39.18 &              50.9 &             69.29 &              83.1 &                   \textbf{22.09} \\

    %& 10 & 4 &  17.65 &   19.61 &   35.29 &   47.06 &  50.41 \\
    %& & & & & & & \\
    %\Xhline{3\arrayrulewidth}
     %& & & & & & & & & & & & \\
    Guidance \textbf{odd} $g$ distance  &      25.41 &       32.17 &        46.8 &       67.31 &             35.88 &                             \textbf{63.06} &             70.65 &             80.94 &             89.16 &                   \textbf{14.01}  \\
    \Xhline{3\arrayrulewidth}
    \end{tabular}
    }
    \caption{Aggregated short-term planning evaluation for odd (unveiling utterances of the dialogue partner) and even distances (which would be uttered by the transformer itself).}
    \label{tab:STPEvalShort}
    %}
\end{table*}
\subsection{Evaluation \& Discussion} \label{sec:EE}
%First, we start with the Imaginary Embeddings themselves for which we have measured the average cosine similarity distance of utterances within dialogues as shown in figure \ref{fig:AVGSimilarity} on the Dialog Test Corpus. While the model does not hit the original training target similarity in average, we can observe that the model is directional sensitive (see red false direction similarity) as well as distance aware (see green).
In the following sections, we investigate how well these embeddings perform on our introduced \textbf{LSTPE} (§\ref{sec:LSTPE}) and on the next utterance selection task. In the main paper, we focus on our empirical findings and present the results of the experiments for space reasons in aggregated form. We provide a detailed analysis in the appendix, where we explore examples as well as demonstrate the curved property of dialogues in these embeddings. This is illustrated as vector chains in figure \ref{fig:curvedDialogues} or the average similarity of different distances and directions within dialogues (appendix \ref{app:IMEA}).

\subsubsection{Short-Term Planning}
%\begin{table*}[!htbp]

As shown in the short-term planning aggregated results table \ref{tab:STPEvalShort}, we split the results based on odd distance length (unveiling utterances of the dialogue partner) and even distance (which would be uttered by the transformer). Both have at least 20\% of the true candidate utterances in the top 5 (Hits@5) (of 100) ranks, 50\% in the top 25 (Hits@25), and a max average rank of 32.56. We observe that speaker token-based imaginary embeddings on odd distances can even achieve 63\% in the top 5 (Hits@5) with the highest average rank of 14.01. This can be expected as odd utterances will be uttered by our dialogue partner which we can greatly influence by our preceding utterance. Interestingly, we find that it is significantly easier to plan 3 turns ahead rather than 2 turns. This is portrayed in the detailed analysis based on the history length, goal distances, and the first goal distance (goal in distance) in table \ref{tab:STPEval} (appendix). Our analysis unveils that the DailyDialog models have an advantage through their more diverse utterance distribution in selecting the true candidate utterance. Furthermore, they perform more consistently across different history lengths and goal distances. MDC, on the other hand, performs overall better but has a higher variance in its performance (with samples of different history lengths and goal distance). Concluding that the score distribution in the ranking process is either more strongly peaked (most in data sets with lots of request intents) or it more is flattened (especially on data with majorly inform intents). We explore this in detail in the appendix \ref{app:ESTPE}. This flattened score distribution can be expected as in many cases of providing information, the actual information has little impact on future turns considering a structured task-oriented setting (e.g. replying on how many people will attend a reservation).

\subsubsection{Next Utterance Selection based on Curved History}\label{sec:uttSelect}
%\begin{figure*}[hbt!]
\begin{figure*}[ht]

    \centering
    \includegraphics[width=0.75\textwidth]{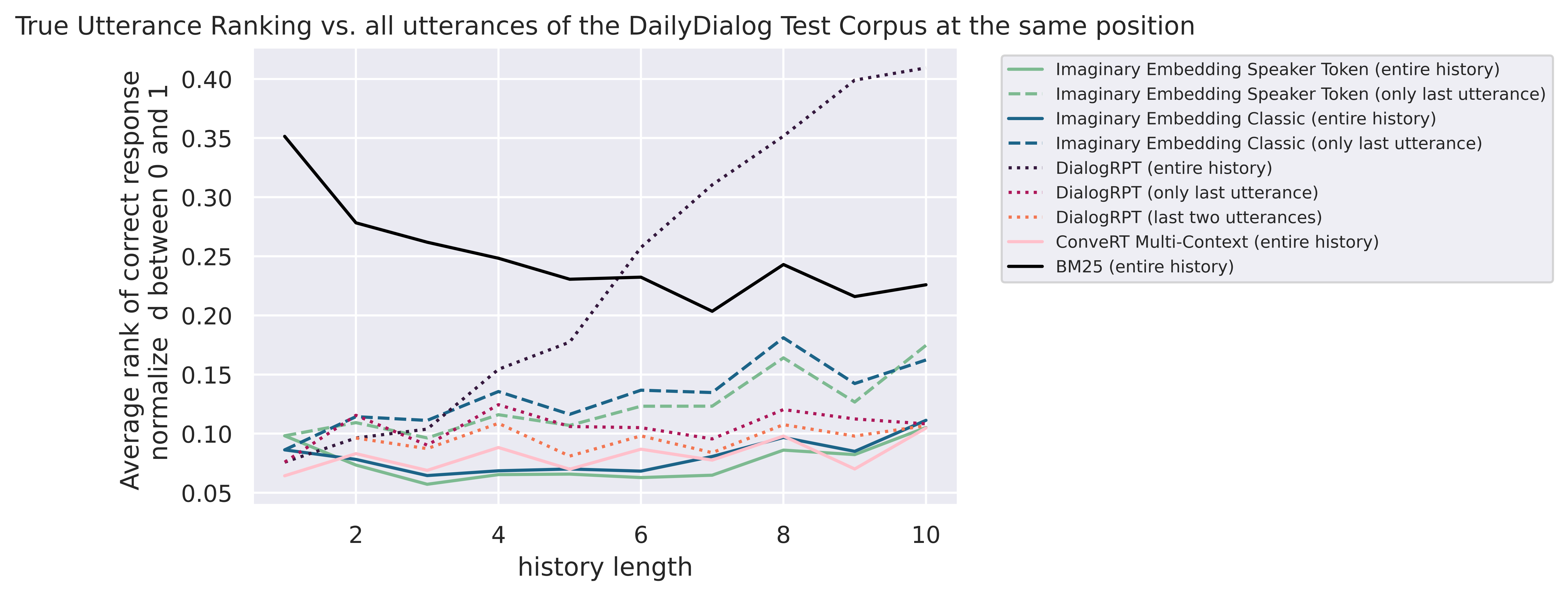}
    \caption{Normalized average rank of next utterance selection based on dialogue history on DailyDialog. Demonstrated are different Curving variants (only the last utterance or the entire history), classic as well as Speaker Token-based embeddings. As baselines, we utilize the pre-trained DialogRPT (human vs random utterance task), the pre-trained ConveRT as well as BM25.}
    \label{fig:curvedSequenceDD}
\end{figure*}
The sequence modeling capability is evaluated based on the normalized average rank (of the true following utterance compared to all other utterances at the same position of the corresponding corpus).
We find that the DailyDialog corpus clearly outperforms MDC across all variations. As we demonstrate in figure \ref{fig:curvedSequenceDD}, DailyDialog performs best with an average rank in the top 10\% over all history lengths (the entire history projected in the curved space with speaker tokens). For sequences longer than 2 turns, it even outperforms all our baselines DialogRPT (human vs. random) by at least 2.8\% and ConvRT by 0.5\%.

Overall, we find that DialogRPT has trouble with increasing sequence lengths as input and find that keeping the last two utterances performs best. Notably, we can reduce the computation costs of the dialogue context compared to DialogRPT and also ConvRT due to our relativistic approach which we explore in more detail in the appendix \ref{sec:speed}.
While our experiments on MDC for the next utterance selection show weak results, in summary, MDC shows the same fluctuations between primarily inform \& requests intents. While the ranking approaches based on only the last utterance are most of the time superior, we observe on odd turns (where we have a lot of request intents) the entire history usually performs better relative to even distances. Conversely, we notice that approaches based on only the last utterance are especially good on turns where we see more informing intents (replying to the request). We further explore this in the appendix \ref{sec:MDCNext}.

\subsubsection{Long Term Planning Evaluation} \label{sec:LTPEE}
\begin{table*}[!htbp]
\centering
 %\medium{
    \renewcommand{\arraystretch}{1.0}
    
    \centering
    
    \resizebox{\textwidth}{!}{
    
    \begin{tabular}{l | c c c c c c | c c c c c c}
    %\Xhline{3\arrayrulewidth}
     & \multicolumn{5}{c}{}{\textbf{Imaginary Embedding w.o. Speaker Token}}  & &  \multicolumn{5}{c}{}{\textbf{Imaginary Embedding with Speaker Token}}  \\

      & \multicolumn{3}{l}{}{\textbf{partially ordered}} & & \textbf{\thead{Reverse \\ order}} & & \multicolumn{3}{l}{}{\textbf{partially ordered}} &  & \textbf{\thead{Reverse \\ order}} \\
     
    %\Xhline{3\arrayrulewidth}
    \multirow{2}{*}{\textbf{Model}}& 
    \multirow{2}{*}{\textbf{\thead{Hits@1 \\ (in \%)}}} & 
    \multirow{2}{*}{\textbf{\thead{Hits@2 \\ (in \%)}}} & 
    \multirow{2}{*}{\textbf{\thead{Hits@3 \\ (in \%)}}} & 
    \multirow{2}{*}{\textbf{\thead{Hits@4 \\ (in \%)}}} &
    \multirow{2}{*}{\textbf{\thead{Hits@1 \\ (in \%)}}} &
    \multirow{2}{*}{\textbf{\thead{Average\\ Rank}}} &
    \multirow{2}{*}{\textbf{\thead{Hits@1 \\ (in \%)}}} & 
    \multirow{2}{*}{\textbf{\thead{Hits@2 \\ (in \%)}}} & 
    \multirow{2}{*}{\textbf{\thead{Hits@3 \\ (in \%)}}} & 
    \multirow{2}{*}{\textbf{\thead{Hits@4 \\ (in \%)}}} &
    \multirow{2}{*}{\textbf{\thead{Hits@1 \\ (in \%)}}} &
    \multirow{2}{*}{\textbf{\thead{Average\\ Rank}}} \\
    & & & & & & & & & & & & \\
     & & & & & & & & & & & & \\
    \Xhline{3\arrayrulewidth}
    %\Xhline{3\arrayrulewidth}
    \multicolumn{8}{l}{\textbf{DailyDialog Test Corpus }} \\
     %\textbf{DailyDialogue Test Corpus} & & & & & & & \\
         %& & & & & & & & & &\\
             IEC &                       49.99 &                       70.62 &                       85.26 &                       93.42 &                    79.17 &               2.01    &                             51.60 &                             72.22 &                             86.82 &                             94.94 &                          81.18 &                    1.94   \\
     %& & & & & & & & & & & & \\
   IEC \& CU &                       50.69 &                       71.24 &                       85.09 &                       93.63 &                    78.54 &              1.99 &                             51.07 &                             72.98 &                              86.9 &                             94.97 &                          79.87 &                    1.94 \\
    %& & & & & & & & & & & & \\
     GC &                       57.87 &                       82.47 &                         - &                         - &                      - &               1.6 &                             57.32 &                             83.89 &                               - &                               - &                            - &                    1.59 \\
     
    \Xhline{3\arrayrulewidth}
     \multicolumn{8}{l}{\textbf{MDC Test Corpus}} \\
    %\textbf{MDC Test Corpus} & & & & & & & \\
     %& & & & & & & \\
     IEC &                       58.72 &                       77.43 &                       90.28 &                       96.38 &                    85.28 &              1.77 &                             56.83 &                             77.50 &                             90.19 &                             95.44 &                          84.52 &                    1.80 \\
     %& & & & & & & & & \\
    IEC \& CU &                       61.59 &                       77.72 &                       90.15 &                       96.79 &                    86.25 &              1.74 &                             58.63 &                             78.62 &                             91.20 &                             95.72 &                          85.44 &                    1.76 \\
     %& & & & & & & & & \\
     GC &                       66.30 &                       89.61 &                         - &                         - &                      - &              1.44 &                             56.05 &                             80.59 &                               - &                               - &                            - &                    1.64 \\
    \Xhline{3\arrayrulewidth}
    \end{tabular}
    }
    \caption{Aggregated Long-Term Planning Evaluation on 3 goals with ((2, 2, 2), (2, 2, 0) and (2, 2, 1)) with (history length, goal distances, first goal \textbf{in} distance). Models include Imaginary Embedding Chain (IEC),  Imaginary Embedding Chain + Curving (IEC \& CU), and Greedy Curving (GC).}
    \label{tab:LTPEvalShort}
    %}
\end{table*}
% describe finding and then refer to 

%old:
The short turn length of the two corpora becomes especially troublesome in the long-term planning evaluation. Here, we are limited to short context/history lengths as well as short goal distances and (first) goal \textbf{in} distances.%We provide all these data sets in GITHUB with a detailed description. 
Across all models and datasets, we observe a solid average rank of 1.87 (between 1 and 2 for all approaches) on identifying the correct order of $3$ goal utterances within their 6 possible orders as table \ref{tab:LTPEvalShort} unveils.  Note that Greedy Curving has only to predict only the immediate next goal (1/3) while the other LTP models the entire order (1/6). While our MDC embeddings had especially trouble with utterance selection in width (selecting an utterance from the same dialog depth §\ref{sec:uttSelect}), we find that MDC shows a stronger performance on greedy goal selection (Greedy Curving (GC)) on classic embeddings thanks to the solidified sequential structure of task-oriented dialogues. This advantage lets MDC outperform DailyDialog also on all other approaches. When Speaker tokens come into play, however, MDC drops while DailyDialog improves in performance compared to classic imaginary embeddings. Imaginary Embedding Chains (IEC) and with curved context (IEC \& CU) show similar performance in aggregated form. However, when the context is close (i.e. the first goal is not far away) IECs with a curved context prevail. This changes with increasing distance of goals or first goal \textbf{in} distance as highlighted in table \ref{tab:LTPEval} of the appendix. Here, IECs with no context keep an advantage. Similarly, we observe a drop in performance over longer distances for Greedy Curving. In terms of the MDC planning capability, the performance drop-off between the two most common intents, request and inform, is similar, although not as severe as in short-term planning or the next utterance selection. %We provide a detailed analysis of the robustness of these embedding chains in the 

\section{Conclusion}\label{sec:conclusion}
In this paper, we introduced Curved Contrastive Learning, a novel technique for generating forward-entailing language embeddings. We demonstrated that these can be utilized on various sequence modeling tasks by only using the cosine similarity between the separately encoded sequence members in the curved space. In particular, for the next utterance selection by imagining the closeness of every context utterance to candidate utterances in the curved space (where DailyDialog's true utterances are constantly in the top 10\%), outperforming our pre-trained baselines DialogRPT and ConvRT on sequences longer than 2 turns while reducing encoding costs. Furthermore, we have shown their pattern recognition ability on the ordering/identification of future representations (with an average rank of 1.87/6) \ even at longer distances and far apart. We also demonstrated that these embeddings can be applied to guiding dialogue transformers to approach a goal over multiple turns. In particular, by imagining the closeness of candidate utterances towards the goal through the transitive properties of the curved space. Following up on our claim, that even chit-chat can be considered goal-oriented \textbf{(RQ1)}, we find strong evidence of planning capability in chit-chat conversations over multiple turns. E.g. 48.83\% /  61.56\%  (within the top 5 / top 10 utterances in the re-ranking) on 3 turns ahead. Our \textbf{RQ2} can be answered by the fact that we observe significant differences in the plannability of different intents. Our empirical analysis shows that request intents are significantly easier to plan than informing intents.
%We provide the first analysis of these embeddings regarding the plannability of dialogues, where we find strong fluctuations between different intents \textbf{(RQ2)}. %As we acknowledge in limitations §\ref{app:limit},
While our focus in this paper was mainly on the introduction of Imaginary Embeddings and their utilization to dialogue planning, we leave much more space for further evaluation, analysis, and applications on the curved properties of our \cancel{universe} \footnote{In tribute to our fellow researchers in the field of physics for their inspiring work on the curvature of spacetime} embeddings in future works. % Reality is curved, we argue NLP should be too!

\section{Limitations}\label{app:limit}
One of our limitations is that the data is split for short-term planning and long-term planning at fixed positions which on one side shows the overall planning capability on different datasets unbiasedly but on the other hand mixes the planning ability of the datasets with the overall performance of the embeddings. We have demonstrated in section \ref{app:exampleSHORT} that this can lead in many cases to unplannable examples. While this means that our embeddings should overall perform better than our results suggest, in the future, we should create either a human-filtered dataset where planning is always possible or either create a human benchmark as a further baseline. Furthermore, we rely in short-term planning (transformer guidance) on the generated utterance distributions by transformers where we have to balance between semantic diversity and the likelihood of utterances.  We control these with temperature and nucleus sampling (top p) and found the best trade-off with a temperature of $0.8$ and a top p of $0.8$. Nonetheless, this can still lead to utterances that might lead to the goal but that would be not considered by humans as very likely based on the given context as we explore in \ref{app:exampleSHORT}. Furthermore, in the next utterance selection, we utilize the publicly available checkpoints which have been evaluated in the paper \citep{DBLP:journals/corr/abs-2009-06978} on DailyDialog but both were seemingly not trained on an MDC-like task-oriented corpus. 
% Should we add 
Since we find that the next utterance selection based on the curved property of the context in a task-oriented setting like MDC is almost always worse than just taking the last utterance, we have not expanded experiments in this domain. 

\section{Ethics}\label{app:ethics}
Like other language models, our model is prone to bias from training data sets \citep{Schramowski2022}\citep{DBLP:journals/corr/abs-1908-09635}. This is something to keep in mind when fine-tuning the model for domain adaptation. Since the models are for guidance only, we do not see any direct threats related to language generation. Still, if an individual intentionally wants to harm others and trains a language model to generate harmful utterances, our model could be employed to support this process. In contrast, however, we argue that these embeddings have great potential through their transitive properties to foresee and deflect harmful utterances from afar. Considering the risk that language models pose to humans \citep{DBLP:journals/corr/abs-2112-04359}, these embeddings could be utilized as a filter on top of generative language models, e.g. removing utterances that would increase the probability of leading to an utterance of a large set of harmful utterances. Our proposed model has a relatively small model size and shows higher efficiency during training \& inference compared to DialogRPT and ConvRT, therefore we see great potential for reducing the carbon footprint in utterance retrieval tasks, in accordance with recent efforts in NLP \citep{DBLP:journals/corr/abs-1906-02243} \citep{DBLP:journals/corr/abs-2104-10350}.

%allow us to greedily move through a conversational curved semantic space for short-term planning. We find strong evidence on short ter

%At the beginning of this paper, we raised the claim, motivated based on existing psychology literature, that even chit-chat conversations can be considered as goal-oriented. Based on our detailed analysis of short-term planning in \ref{tab:STPEval}, we find an average planning capability of  on our chit-chat data set DialyDialog. Whether the individuals had the intent to push the conversations over 3 turns toward the corresponding goal remains unanswered but we demonstrated that we can guide the conversation in the majority. Furthermore, it remains unanswered why there is such a significant drop between unveiling utterances in dialog partners over longer distances than even ones that are closer. 

%We demonstrated that we can utilize the Imaginary Embeddings for a light weighted sequence modeling with. 

%Find evidence for majorly planning with (2,3) 52.53\% in top 5  planning capability in Daily Dialog 

% end sentence:
%As Albert Einstein showed in his paper in 1914, that reality is curved, we demonstrated today 

%- knowledge graph biased distribution prediction with a combination of rules --> only problem divergent utterances

%--> zero shot 
%--> Prototypes psychology entail 
%--> identify entailing/contradicting utterances based on history 

%\bibliography{custom}
%\bibliography{custom}
\bibliographystyle{acl_natbib}

\newpage

\appendix
 
%In particular, for short-term planning, we  rely on the utterance generation of transformers for to find a good line between diversity and likelihood of utterances

%that might lack diversity in the utterance generation process which can affect the results negatively as well as positively. We - Though our models show strong results, we have grounded all of our evaluations based on test corpora, splitting the goal ordering / approaching based on a fixed history length, goal distances, and (first) goal in distance without considering the human plan ability given the dialog snippets . This might make our models look worse than they actually are. As described in the evaluation one drawback remains the lack of long dialogues which demonstrates a drop-off in performance.

%he diverse distribution of ImagineGPT, is important for guidance. 
%Performance next utterance selection. More detailed analysis needed --> paper focuses on the introduction of the new concepts. 
\section{Attribution}
This work stems from the mandatory master's internship of Justus-Jonas Erker at the German Research Center for Artificial Intelligence supervised by Stefan Schaffer and Gerasimos Spanakis.

\section{Imaginary Embedding extended analysis}\label{app:IMEA}
\begin{figure}[ht]
    \centering
    \includegraphics[width=0.45\textwidth]{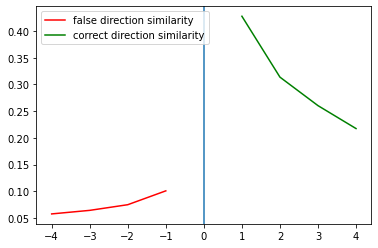}
    \caption{Average Imaginary Embedding Similarity to correct and false direction utterances based on turn distance on DailyDialog Test Corpus}
    \label{fig:AVGSimilarity}
\end{figure}

\begin{figure}[ht]
    \centering
    \includegraphics[width=0.45\textwidth]{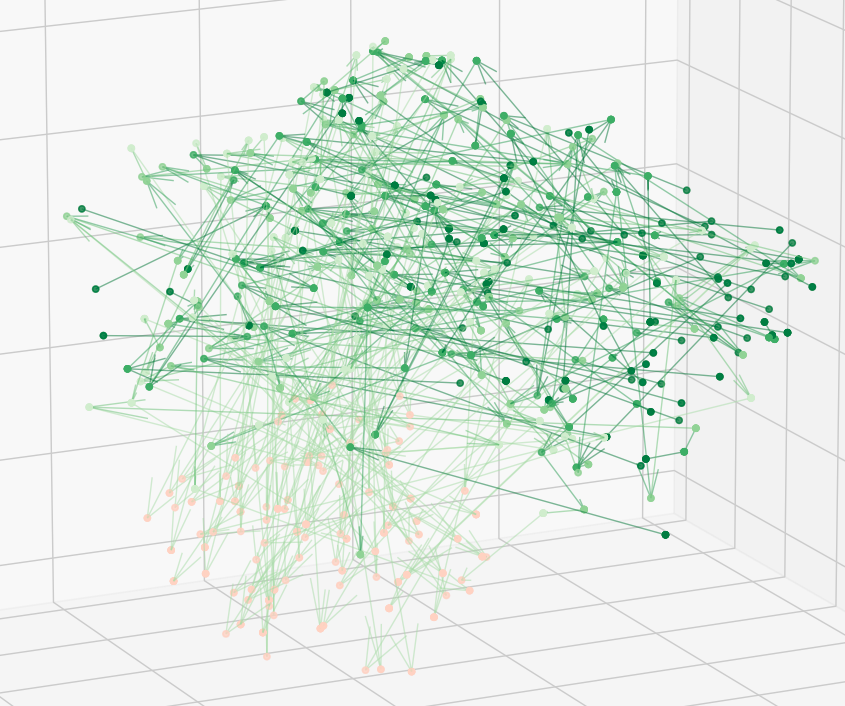}
    \caption{t-SNE visualization of first 4 utterances of the first 100 dialogues of the DailyDialog Test Corpus in curved Embedding Space.  From Dark green to light green ($u_1 \rightarrow u_2 \rightarrow u_3$) nodes as well as edges encoded with the \texttt{[BEFORE]} token to $u_4$ encoded with \texttt{[AFTER]} token as light red.}
    \label{fig:curvedDialogues}
\end{figure}

We analyze the Imaginary Embeddings based on their average similarity to different distances of utterances pairs within dialogues as well as their direction as shown in figure \ref{fig:AVGSimilarity}. While the model's average similarity is far from the training objective, the scores show a favorable decay considering the distance for positive examples as well as a relatively low similarity for false direction utterance pairs.
Furthermore, we have illustrated the curved property of these embeddings as directed graphs of dialogues in figure \ref{fig:curvedDialogues} where we notice a tendency of utterances at the beginning of the dialogue in the close right and the last utterance (encoded with the after token) deeper on the left.

\section{Next Utterance Selection Extended Analysis}
For the next utterance selection we provide an extended description for our speed comparison as well as the MDC results.
\subsection{Computation Comparison}\label{sec:speed}

% Computation results
%INFO:root:ConvRT Transformer Encodings for Candidates: 6219
%INFO:root:ConvRT Transformer Encodings costs History: 6219
%INFO:root:ConvRT Transformer Encodings costs History for every Utterance: 26733x
%INFO:root:ConvRT Transformer Encodings costs Total: 12438

%INFO:root:Imaginary Embedding Transformer Encodings for Candidates: 7740
%INFO:root:Imaginary Embedding Transformer Encodings costs History: 7740
%INFO:root:Imaginary Embedding Transformer Encodings costs Total: 15480
%INFO:root:Imaginary Embedding Transformer Encodings time in seconds for Candidates: 3.778195
%INFO:root:Imaginary Embedding Transformer Encodings time in seconds for History: 3.109647
%INFO:root:Imaginary Embedding Transformer Encodings time in seconds Total: 6.887842

%
%
Since the bi-encoder architectures are significantly more efficient than DialogRPT, we compare ConvRT and Imaginary Embeddings in more detail. Considering the encoding of utterances for some sequence of length $n$, ConvRT requires each context representation to encode every previous utterance again with $\mathcal{O}(n)$ while Imaginary Embeddings only encodes the last utterance $\mathcal{O}(1)$. Therefore, the entire next utterance selection task for the DailyDialog Corpus (up to a context length of 10 utterances) requires ConvRT to generate 6219 context representations where in total 26733 utterances are encoded. Imaginary Embeddings reduce the computation of encoding utterances through its relativistic approach by a factor of $\frac{26733}{6219}=4.3$. In terms of context to candidate utterance matching, Imaginary Embeddings can pre-compute the entire context until utterance $n-1$ with $(batch, h\_len-1, emb)$ while the dialogue partner is speaking. Since we got normalized embeddings from the sentence transformer we can compute the cosine similarity-based score for context and candidate pairs in one simple batch matrix multiplication $U \odot H.T$ by transposing the history with dimensions $(1,2)$. Following we sum across the second dimension (history dim) like equation \ref{eq:hischain} illustrates and store the score matrix $s_{1,...,n-1}$ in memory. At inference, we have to compute scores only between the last utterances and the candidate utterances matching the number of dot products with ConvRT. Once the new score matrix $s_n$ for every pair is generated we simply sum the two score matrices $s = s_{1,...,n-1} + s_n$. 

%In this section, we will investigate the speed of our introduced curving technique to rank 1000 incoming utterances for 1000 different contexts by comparing it to DialogRPT.  Thanks to the curved property we have to only encode the new incoming utterances with the after token which takes on GPU (\texttt{A100-40GB}) around one second (1.16s). Following, we load the tensors on GPU, in particular, the history tensor $H$ (encoded with the after tokens) with $(batch, h\_len, emb)$ (here with a batch size of 1000 and a history length of 3) as well as the following utterance candidates $U$(with $(batch, n\_cand, emb)$ which takes 0.091 seconds. Since we got normalized embeddings from the sentence transformer we can compute the cosine similarity-based score for the $1000 000$ dialogues in \textbf{one} simple batch matrix multiplication $U \odot H.T$ by transposing the history with dimensions $(1,2)$. Following we sum across the second dimension (history dim) like equation \ref{eq:hischain} illustrates. Compared to DialogRPT on the same \texttt{A100-40GB} GPU, these two tensor computations (batch matrix multiplication \& sum through the history dimension) take only $0.000531$ seconds for a history length of 3 while DialogRPT needs for the same task, with a batch size of 32, 4023.78 seconds (67 minutes). As $\frac{4023.78}{0.000531} = 7.58 \cdot 10^6 $ we conclude a 7 million time faster GPU computation time. Thereafter, we sort the similarity scores and search the index of the true utterance and return the corresponding rank which takes around 0.212 seconds.
\subsection{MDC Results}\label{sec:MDCNext}
\begin{figure}[ht]
    \centering
    \includegraphics[width=0.49\textwidth]{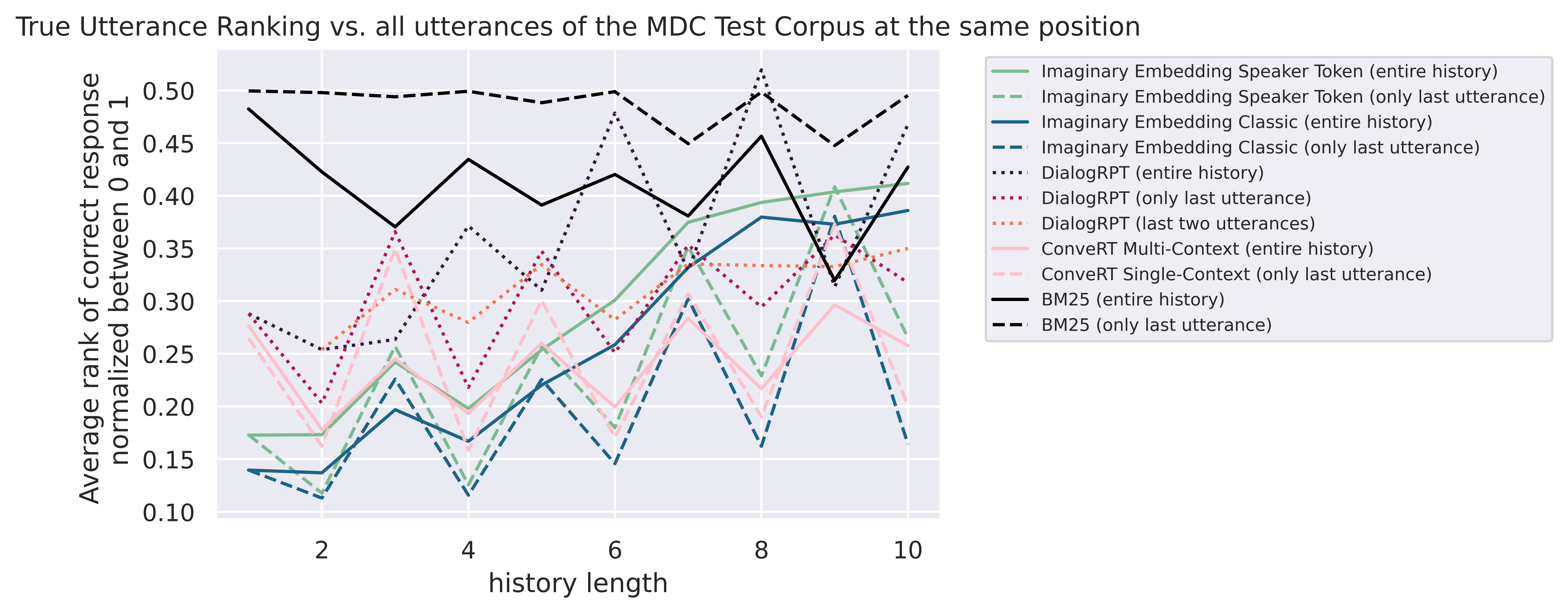}
    \caption{Normalized average rank of next utterance selection based on dialogue history on MDC. Demonstrated are different Curving variants (only the last utterance or the entire history), classic as well as Speaker Token-based embeddings. As baselines, we utilize the pre-trained DialogRPT (human vs random utterance task), the pre-trained ConveRT as well as BM25.}
    \label{fig:curvedSequenceMDC}
\end{figure}

We demonstrate the results of the MDC next utterance selection in figure \ref{fig:curvedSequenceMDC} where we observe as described in the main paper the symmetry between inform and request intents either profiting from only the last utterance or the entire history. 

\section{Speaker Token Learning Objective} \label{sec:STLO}
\newcommand\scalemath[2]{\scalebox{#1}{\mbox{\ensuremath{\displaystyle #2}}}}

\begin{equation}
\label{equation:trainingdataconstruction2}
    \scalemath{0.7}{
    \forall i \in \{1,..,l\}: \begin{cases}
    
  ([E] [B] \ u[0], [A] \ u[i], s= \frac{l-i}{l} & \text{if } i \mod 2 = 0
\\
  ([O] [B] \ u[0], [A] \ u[i], s= \frac{l-i}{l} & \text{if } i \mod 2 \neq 0
\\
  ([E] [B] \ u[i], [A] \ u[0], \  s= 0 & \text{if } i \mod 2 = 0
  \\
  ([O] [B] \ u[i], [A] \ u[0], \  s= 0 & \text{if } i \mod 2 \neq 0
  \\
  ([O] [B] \ u[i], [A] \ u'[r], s= 0 & (p=\frac{1}{4})
   \\
  ([E] [B] \ u[i], [A] \ u'[r], s= 0 & (p=\frac{1}{4})
   \\
  ([O] [B] \ u'[r], [A] \ u[i], s= 0 & (p=\frac{1}{4})
   \\
  ([E] [B] \ u'[r], [A] \ u[i], s= 0 & (p=\frac{1}{4})
\end{cases}
}
\end{equation}

where $[A]$ = \texttt{[AFTER]}, $[B]$ = \texttt{[BEFORE]}, $[E]$ = \texttt{[E]}, $[O]$ = \texttt{[O]}, $u$ the utterances in the observed window, $u'$ a set of random utterances, and $s$ the cosine similarity score. For the random utterance matching we assign an equal probability $p$ to every possible combination.

\section{Extended Short-Term Planning Evaluation} \label{app:ESTPE}
As part of the extended Short Term Planning Evaluation, we investigate the extended results based on the history length, goal distances, and the first goal distance (goal in distance) in table \ref{tab:STPEval} and demonstrate examples. 

\subsection{Detailed Short-Term Planning Evaluation}
Table \ref{tab:STPEval} unveils that additional speaker tokens show improvement in the MDC Test corpus across all tested categories. While classic embeddings show on MDC a similar performance across all even distances, we can observe two spikes at position $(3,1)$ and $(5,1)$ with $(h_l, g_d)$ on odd distances with $51.17$\% / $45.80$\% in the top 5 respectively. At these positions, we monitor a 33\% increase in the standard deviation on average of the distribution of guidance scores i.e. that the model is much more decisive in its ranking. We analyzed the intent at these positions and find a two times increase in requests and a 38\% decrease in inform intents to the data set's average. While the speaker token-based embeddings show that we can overcome this gap for odd distances, we still find that the two lowest performers on $(4,1)$ \& $(4,3)$ with "only" $53.03$\% \& $51.45$\%  in the top 5 have all a minimum of 80\% of informing intents. Since the two corpora use separate latent spaces, we do not compare them on a simple standard deviation. Instead, we take the sum of average standard deviations as a baseline and divide it by the sum of the standard deviations (for each data set) of the standard deviations (for each transformer utterance distribution) to measure the variation in performance over different testing parameters history length, goal distances, (first) goal in distance. With a 35\% higher score, DailyDialog shows less variance through different test parameters. Nonetheless, we find that DailyDialog has a 12\% higher semantic variance across all utterances in the transformer-generated distributions than MDC by measuring their average semantic similarity with a simple semantic sentence transformer.
\subsection{Examples of Short-Term Planning} \label{app:exampleSHORT}
\begin{figure*}[ht]
    \centering
    \includegraphics[width=1.0\textwidth]{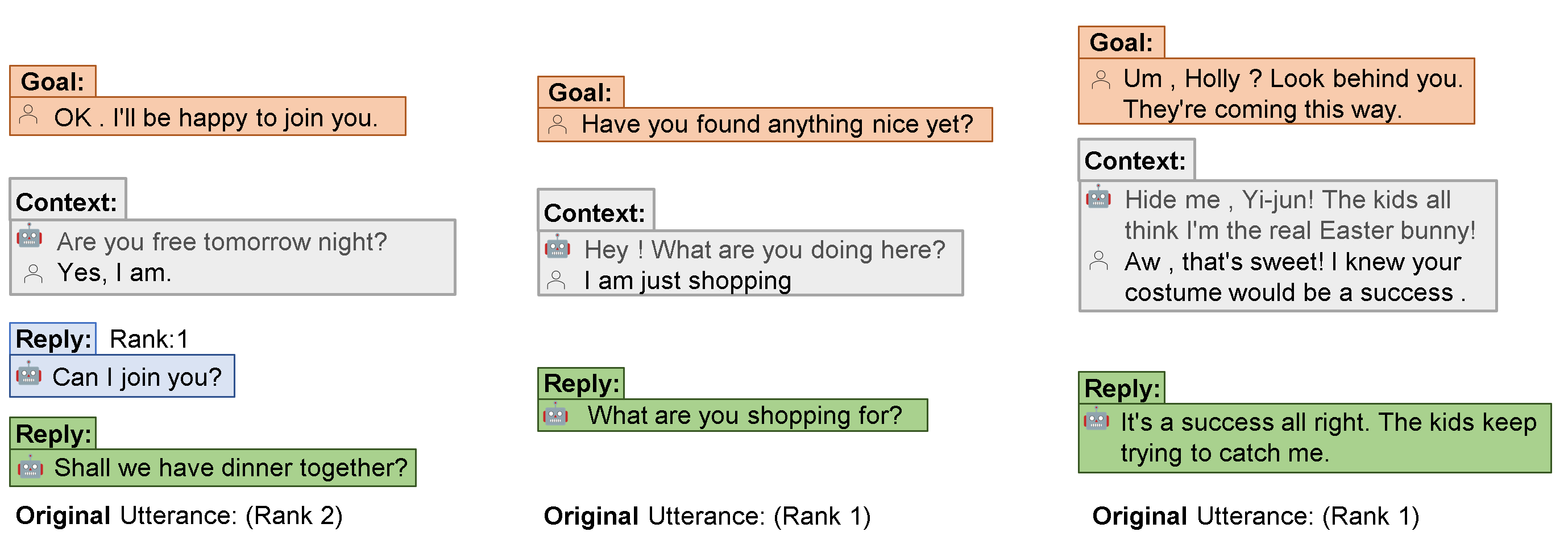}
    \caption{Good Ranking Examples on DailyDialog Test Corpus with a history length of 2 and a goal distance of 3. The goal in red, the context in grey, the true utterance in green, and the transformer-generated utterance in blue}
    \label{fig:GoodRankSTP}
\end{figure*}

\begin{figure*}[ht]
    \centering
    \includegraphics[width=1.0\textwidth]{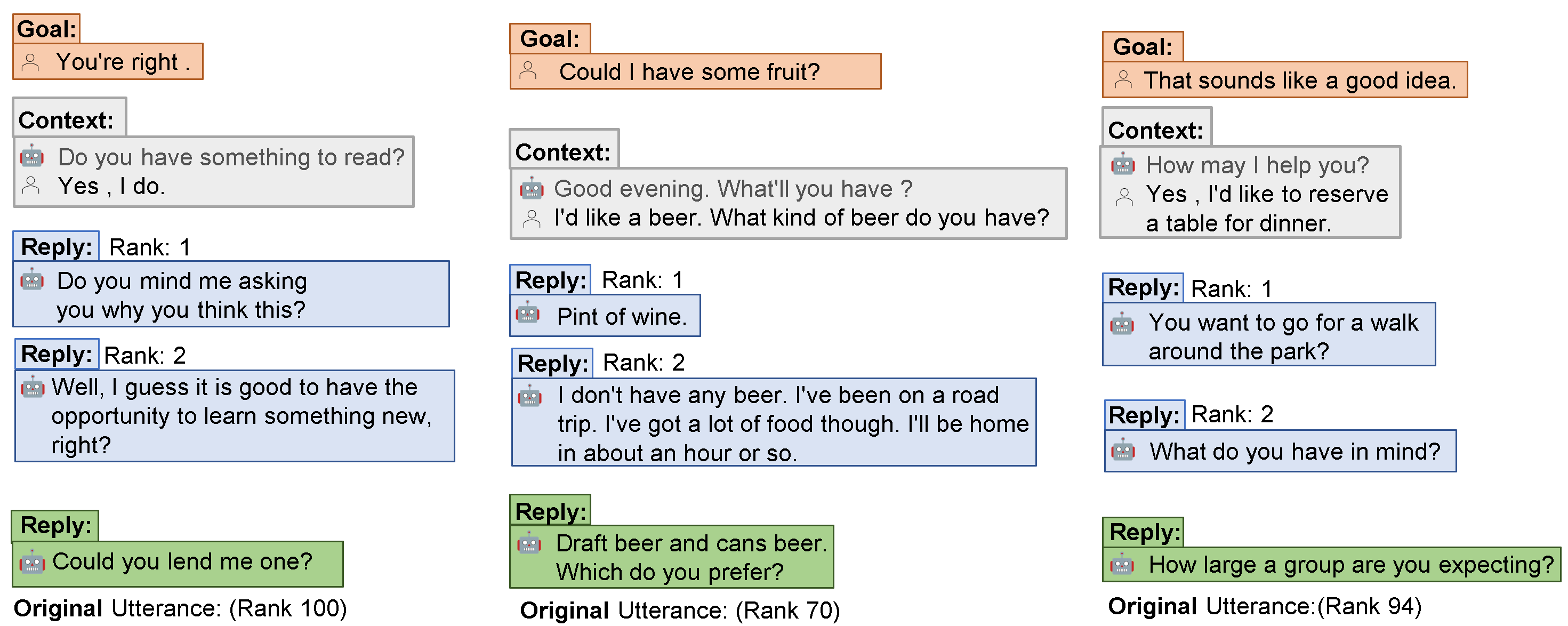}
    \caption{Bad Ranking Examples on DailyDialog Test Corpus with a history length of 2 and the goal distance of 3. The goal in red, the context in grey, the true utterance in green, and the transformer-generated utterance in blue}
    \label{fig:BadRankSTP}
\end{figure*}

While we provide construction of our evaluation datasets, we still want to highlight some of the strengths and weaknesses of our introduced embeddings. In the example on the left of figure \ref{fig:GoodRankSTP}, we can see that without knowing what the person is going to say, the model can sometimes move toward the goal too greedily. In the example on the right, we see that the model can also understand more complex relations, where the only way to get to a conversation state where someone would utter "look behind you. They are coming this way" would be in a manner of playing catch me as the model ranks it on the first position.
A lot of the weaker ranking results are due to the fixed split of data as demonstrated in figure \ref{fig:BadRankSTP}. We observe in the first example (left) that the model tries to unveil the utterance "You're right" by trying to get the other person into an argument (rank 1) where it hopes the person would then agree to their own opinion 3 turns later or by trying to unveil the utterance right away (rank 2). In the example in the middle, we see the drawback of purely relying on the transformer's context-aware utterance generation as the selected utterance of "pint of wine" might be closer to fruits than beer but at the same time is not a valid answer. This can be also observed in the last example (right). 

%\section{Robustness Imainary Embedding Chains}\label{app:robust}
%--> same test needed

\begin{table*}[!htbp]
\centering

 %\medium{
    \renewcommand{\arraystretch}{1.2}
    
    \centering
    %\label{tab:LTPEval}
    \resizebox{\textwidth}{!}{
    
    \begin{tabular}{l c c c | c c c c c | c c c c c }
    \Xhline{3\arrayrulewidth}
    & & & & & & \multicolumn{6}{c}{}{\textbf{\thead{Human Utterance Ranking vs 100 utterances sampled \\ from DialoGPT Large / GODEL Large (p=0.8, t=0.8)}}} & \\

     & & & & & \multicolumn{3}{c}{}{\textbf{\thead{Imaginary Embedding \\ without Speaker Token}}} & & & \multicolumn{3}{c}{}{\textbf{\thead{Imaginary Embedding \\ with Speaker Token}}}   \\
    %\Xhline{3\arrayrulewidth}
    \multirow{2}{*}{\textbf{Embedding Type}}& 
    \multirow{2}{*}{\textbf{ \thead{History \\ Length}}}& 
    \multirow{2}{*}{\textbf{ \thead{Goal \\ Distance}}}& 
     \multirow{2}{*}{\textbf{\thead{$n$}}} &
    \multirow{2}{*}{\textbf{\thead{Hits@5 \\ (in \%)}}} & 
    \multirow{2}{*}{\textbf{\thead{Hits@10 \\ (in \%)}}} & 
    \multirow{2}{*}{\textbf{\thead{Hits@25 \\ (in \%)}}} & 
    \multirow{2}{*}{\textbf{\thead{Hits@50 \\ (in \%)}}} &
    \multirow{2}{*}{\textbf{\thead{Average \\ Rank}}} &
    \multirow{2}{*}{\textbf{\thead{Hits@5 \\ (in \%)}}} & 
    \multirow{2}{*}{\textbf{\thead{Hits@10 \\ (in \%)}}} & 
    \multirow{2}{*}{\textbf{\thead{Hits@25 \\ (in \%)}}} & 
    \multirow{2}{*}{\textbf{\thead{Hits@50 \\ (in \%)}}} &
    \multirow{2}{*}{\textbf{\thead{Average \\ Rank}}} \\
    & & & & & & & & & & & & \\
    \Xhline{3\arrayrulewidth}
    %\Xhline{3\arrayrulewidth}
    %\multicolumn{8}{l}{\textbf{DailyDialogue Test Corpus even goal distance $g_d$ (saying goal by yourself)}} \\
     \textbf{DailyDialog Test Corpus} & & & & & & & & \\
         & & & & & & & & & & & & \\
        \multirow{4}{*}{\shortstack{Guidance with \textbf{even} \\ goal  distance $g_d$  \\(saying goal by yourself)}}
       & 2  & 2 & 741 &     23.08 &       31.44 &       50.74 &       70.31 &             33.65 &            24.70 &             33.87 &             55.06 &             75.57 &                   30.28  \\
&2  & 4 & 534 &      23.03 &       31.65 &       48.13 &       66.85 &             35.61 &            22.10 &             32.02 &             51.31 &             71.72 &                   32.57 \\
&5  & 2 & 479 &      25.05 &       31.52 &       44.47 &       63.47 &             38.03 &            20.88 &             29.23 &             49.69 &             69.52 &                   34.87  \\
&5  & 4 & 323 &      15.79 &       22.60 &       39.01 &       56.66 &             43.02 &            17.65 &             24.15 &             42.11 &             66.25 &                   38.27  \\
&10 & 2 & 102 &      48.04 &       51.96 &       60.78 &       77.45 &             27.18 &            36.27 &             45.10 &             61.76 &             70.59 &                   30.37 \\

    %& 10 & 4 &  17.65 &   19.61 &   35.29 &   47.06 &  50.41 \\
    & & & & & & & \\
    %\Xhline{3\arrayrulewidth}
     & & & & & & & & & & & & \\
    
     \multirow{4}{*}{\shortstack{Guidance with \textbf{odd} \\ goal  distance $g_d$  \\(unveiling goal utterance \\ in dialogue partner)}}
  & 2  & 1 & 918 &      42.37 &       50.54 &       66.88 &       84.64 &             21.74 &            70.59 &             78.54 &             87.15 &             94.55 &                    9.15 \\
& 2   & 3 &  651 &    23.66 &       33.33 &       51.00 &       71.89 &             32.05 &            52.53 &             60.52 &             74.04 &             84.79 &                   19.19 \\
& 5  & 1 &  534 &    35.02 &       43.26 &       58.61 &       76.40 &             27.90 &            67.79 &             77.53 &             86.70 &             93.26 &                   10.46 \\
& 5   & 3 & 385 &      18.44 &       23.64 &       40.00 &       61.04 &             40.92 &            48.83 &             61.56 &             76.62 &             85.97 &                   18.83 \\
& 10 & 1 & 183 &      36.61 &       44.81 &       54.10 &       69.95 &             30.49 &            77.60 &             82.51 &             91.26 &             96.72 &                    6.86 \\

    & & & & & & & \\

    \Xhline{3\arrayrulewidth}

    \textbf{MDC Test Corpus} & & & & & & & \\
         & & & & & & & & & & & & \\
        \multirow{4}{*}{\shortstack{Guidance with \textbf{even} \\ goal  distance $g_d$  \\(saying goal by yourself)}} &2  & 2 & 600 &      20.67 &       28.83 &       43.00 &       64.33 &             37.68 &            45.83 &             55.00 &             69.33 &             84.33 &                   20.41 \\
 &2   & 4 & 417 &     21.58 &       31.18 &       47.00 &       67.63 &             36.02 &            47.48 &             55.16 &             70.26 &             83.45 &                   20.85 \\
&3  & 2 & 545 &      22.02 &       32.66 &       50.64 &       69.72 &             33.33 &            34.68 &             44.40 &             66.24 &             78.35 &                   25.08 \\
&3    & 4 & 344 &     26.16 &       38.08 &       53.49 &       77.62 &             28.97 &            41.28 &             53.20 &             67.44 &             85.76 &                   20.93 \\
& 4  & 2 &  417 &    20.62 &       29.50 &       46.28 &       64.99 &             36.58 &            37.89 &             47.96 &             67.63 &             85.13 &                   21.06 \\
& 4   & 4 &  234 &    16.67 &       23.08 &       47.01 &       70.51 &             37.24 &            40.60 &             53.42 &             73.93 &             89.74 &                   18.04 \\
&5  & 2 & 344 &     18.02 &       24.42 &       40.70 &       60.47 &             40.94 &            29.36 &             41.86 &             61.05 &             77.03 &                   26.79 \\
&5   & 4 & 161 &     20.50 &       34.78 &       56.52 &       78.26 &             28.09 &            44.72 &             58.39 &             75.78 &             88.82 &                   17.32 \\
%&10 & 2 &      22.86 &       25.71 &       45.71 &       80.00 &             32.14 &            20.00 &             37.14 &             60.00 &             77.14 &                   30.94 \\

    %& 10 & 4 &  17.65 &   19.61 &   35.29 &   47.06 &  50.41 \\
    & & & & & & & \\
    %\Xhline{3\arrayrulewidth}
     & & & & & & & & & & & & \\
    
     \multirow{4}{*}{\shortstack{Guidance with \textbf{odd} \\ goal  distance $g_d$  \\(unveiling goal utterance \\ in dialogue partner)}}
&2  & 1 &  893 &    20.83 &       27.32 &       40.54 &       61.59 &             38.89 &            63.83 &             69.99 &             81.41 &             90.26 &                   13.46 \\
&2   & 3 & 545 &     31.19 &       38.53 &       55.41 &       73.76 &             29.92 &            69.91 &             77.06 &             83.30 &             90.28 &                   11.78 \\
&3  & 1 & 600 &     \textbf{51.17} &       58.00 &       70.33 &       82.00 &             20.75 &            69.17 &             74.17 &             83.33 &             91.50 &                   12.03 \\
&3    & 3 &  417 &    15.83 &       25.18 &       43.88 &       68.35 &             37.87 &            67.39 &             73.62 &             83.93 &             93.29 &                   11.25 \\
&4  & 1 & 545 &     18.17 &       26.06 &       43.30 &       67.16 &             36.16 &            53.03 &             63.49 &             76.70 &             84.04 &                   18.23 \\
&4   & 3 & 344 &     17.44 &       25.58 &       42.44 &       61.34 &             39.51 &            51.45 &             62.50 &             76.16 &             83.14 &                   18.42 \\
&5  & 1 & 417 &      \textbf{45.80} &       52.28 &       63.07 &       74.34 &             26.56 &            73.38 &             77.22 &             85.85 &             91.85 &                   10.85 \\
&5   & 3 &  234 &     16.24 &       19.23 &       32.91 &       58.55 &             46.47 &            71.37 &             77.78 &             88.46 &             92.74 &                    9.92 \\
%&10 & 1 &      12.00 &       17.33 &       29.33 &       58.67 &             46.79 &            48.00 &             60.00 &             69.33 &             85.33 &                   20.12 \\

     & & & & & & & \\
    \Xhline{3\arrayrulewidth}
    \end{tabular}
    }
    \caption{Detailed Short-Term Planning Evaluation with $n$ (number of evaluation samples)}
    \label{tab:STPEval}
    %}
\end{table*}
\newpage
\section{Long-Term Planning Results}
We present our detailed Long Term planning results in table \ref{tab:LTPEval} as well as examples in the following subsection.
 %\todo{add average rank to LTP? is it okay to have the results in the appendix, should there be an aggregated results table in the paper?}

\subsection{Long-Term Planning Examples}
\begin{figure*}[ht]
    \centering
    \includegraphics[width=1.0\textwidth]{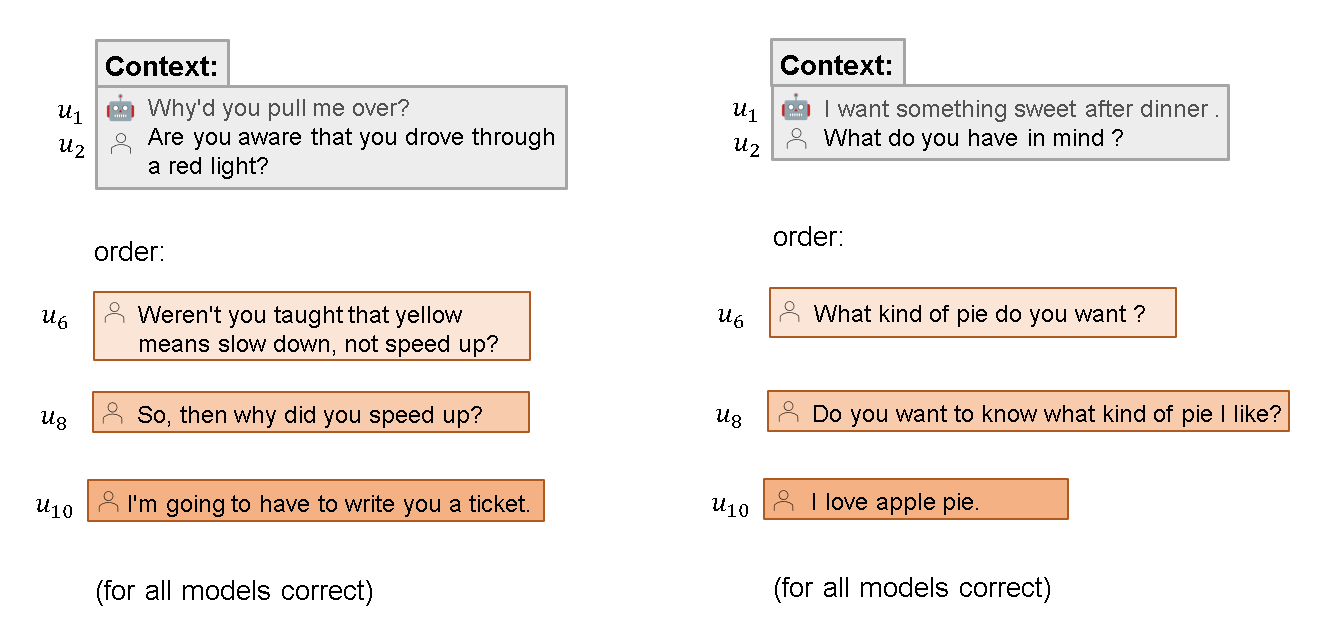}
    \caption{Good Ranking Examples on DailyDialog Test Corpus with history length of 2, the goal distance of 2, and goal in distance of 3}
    \label{fig:GoodRankLTP}
\end{figure*}

\begin{figure*}[ht]
    \centering
    \includegraphics[width=1.0\textwidth]{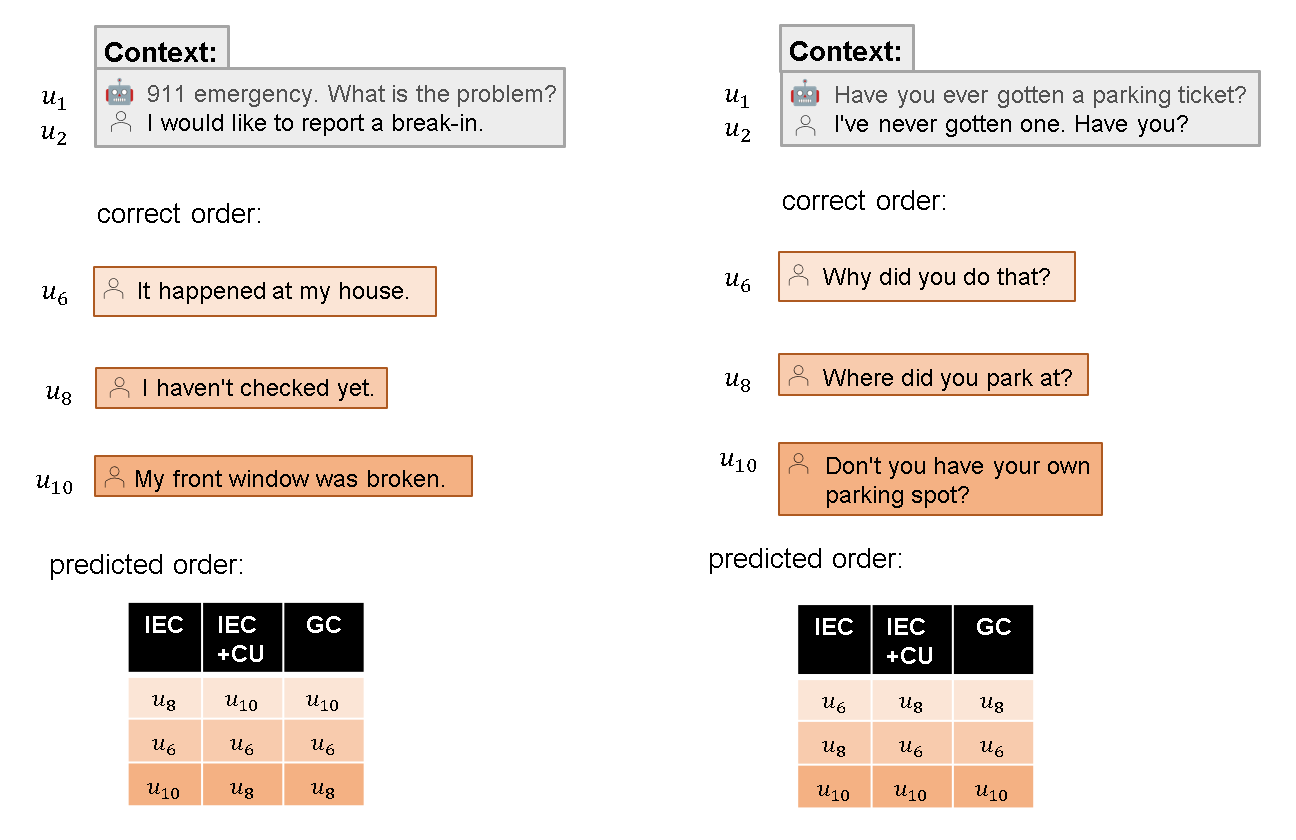}
    \caption{Bad Ranking Examples on DailyDialog Test Corpus with history length of 2, the goal distance of 2, and goal in distance of 3}
    \label{fig:BadRankLTP}
\end{figure*}

Alike for short-term planning, we demonstrate examples to present the weaknesses and as well as strengths of the embeddings. In figure \ref{fig:GoodRankLTP} we show two very easy examples, where we can follow the conversation well without knowing the replies of the other dialogue partner. This changes especially in figure \ref{fig:BadRankLTP} where in the left example it is also for us very difficult to order the corresponding utterances. While one could argue that emergency calls tend to start with the location of the incident, the utterance "I haven't checked yet" makes the ordering of the utterances without any further context very difficult. This can also be observed in the right example of figure \ref{fig:BadRankLTP}, however, one could argue that based on the context to which both IEC+CU and GC have access, the predicted order (of these two) makes more sense than the original reply order. Nonetheless, both examples show that some of these orders are debatable.


\begin{table*}[!htbp]
\centering
 %\medium{
    \renewcommand{\arraystretch}{1.2}
    
    \centering
    %\label{tab:LTPEval}
    \resizebox{\textwidth}{!}{
    
    \begin{tabular}{l c c c c | c c c c  c c | c c c c c c}
    \Xhline{3\arrayrulewidth}
    & & & & & & & \multicolumn{6}{c}{}{\textbf{\thead{LTP Planning Evaluation for 3 Goals}}} \\

     & & & & & &\multicolumn{5}{c}{}{\textbf{\thead{Imaginary Embedding \\ without Speaker Token}}}  & & \multicolumn{3}{c}{}{\textbf{\thead{Imaginary Embedding \\ with Speaker Token}}}  \\

     & & & & & \multicolumn{3}{l}{}{\textbf{partially ordered}} & & \textbf{\thead{Reverse \\ order}} & & \multicolumn{3}{l}{}{\textbf{partially ordered}} &  & \textbf{\thead{Reverse \\ order}} \\
     
    %\Xhline{3\arrayrulewidth}
    \multirow{2}{*}{\textbf{Model}}& 
    \multirow{2}{*}{\textbf{ \thead{History \\ Length}}}& 
    \multirow{2}{*}{\textbf{ \thead{Goal \\ Distances}}}& 
    \multirow{2}{*}{\textbf{ \thead{First Goal \\ In Distance}}}& 
     \multirow{2}{*}{\textbf{\thead{$n$}}} &
    \multirow{2}{*}{\textbf{\thead{Hits@1 \\ (in \%)}}} & 
    \multirow{2}{*}{\textbf{\thead{Hits@2 \\ (in \%)}}} & 
    \multirow{2}{*}{\textbf{\thead{Hits@3 \\ (in \%)}}} & 
    \multirow{2}{*}{\textbf{\thead{Hits@4 \\ (in \%)}}} &
    \multirow{2}{*}{\textbf{\thead{Hits@1 \\ (in \%)}}} &
    \multirow{2}{*}{\textbf{\thead{Average\\ Rank}}} &
    \multirow{2}{*}{\textbf{\thead{Hits@1 \\ (in \%)}}} & 
    \multirow{2}{*}{\textbf{\thead{Hits@2 \\ (in \%)}}} & 
    \multirow{2}{*}{\textbf{\thead{Hits@3 \\ (in \%)}}} & 
    \multirow{2}{*}{\textbf{\thead{Hits@4 \\ (in \%)}}} &
    \multirow{2}{*}{\textbf{\thead{Hits@1 \\ (in \%)}}} &
    \multirow{2}{*}{\textbf{\thead{Average\\ Rank}}} \\
    & & & & & & & & & & & & & \\
    \Xhline{3\arrayrulewidth}
    %\Xhline{3\arrayrulewidth}
    \multicolumn{8}{l}{\textbf{DailyDialog Test Corpus}} \\
     %\textbf{DailyDialogue Test Corpus} & & & & & & & \\
         & & & & & & & & & & & & & \\
         \multirow{7}{*}{IEC} 
          &2 &              2 &                 0 &    385 &                   57.66 &                       72.47 &                       87.79 &                       93.25 &                    81.56 &              1.88 &                             58.70 &                             76.10 &                             91.43 &                             97.14 &                          84.16 &                    1.76 \\
        & 2 &              2 &                 1 &  323 &                     46.13 &                       68.73 &                       84.83 &                       94.74 &                    79.26 &              2.06 &                             51.39 &                             70.90 &                             86.07 &                             94.43 &                          81.42 &                    1.97 \\
        &  2 &              2 &                 2 &       230 &                46.52 &                       67.83 &                       83.04 &                       90.87 &                    74.78 &              2.11 &                             46.96 &                             67.39 &                             83.91 &                             92.17 &                          78.70 &                    2.09 \\

        & 2 &              2 &                 3 &     183 &                  \textbf{44.26} &                       66.12 &                       77.60 &                       91.26 &                    73.22 &              2.20 &                             50.82 &                             68.85 &                             83.61 &                             93.99 &                          77.60 &                    2.03 \\

        &  4 &              2 &                 0 &    230 &                   46.52 &                       67.83 &                       83.04 &                       90.87 &                    74.78 &              2.11 &                             46.96 &                             67.39 &                             83.91 &                             92.17 &                          78.70 &                    2.09 \\

        &4 &              2 &                 1 &    183 &                   44.26 &                       66.12 &                       77.60 &                       91.26 &                    73.22 &             2.20 &                             50.82 &                             68.85 &                             83.61 &                             93.99 &                          77.60 &                    2.02 \\

        &  4 &              2 &                 2 &       102 &                \textbf{43.14} &                       68.63 &                       82.35 &                       92.16 &                    73.53 &              2.13 &                             39.22 &                             60.78 &                             79.41 &                             93.14 &                          64.71 &                    2.27 \\

       & 2 &              4 &                 0 &    51 &                   37.25 &                       56.86 &                       78.43 &                       86.27 &                    76.47 &              2.41 &                             47.06 &                             66.67 &                             96.08 &                             98.04 &                          84.31 &                    1.92  \\
     & & & & & & & & & & & & \\
    \multirow{7}{*}{IEC \& CU} 
    &  2 &              2 &                 0 &   385 &                      57.92 &                       76.36 &                       90.13 &                       95.84 &                    84.42 &              1.78 &                             \textbf{59.74} &                             77.92 &                             92.21 &                             98.18 &                          85.45 &                    1.72 \\

    &    2 &              2 &                 1 &  323 &                     46.75 &                       67.80 &                       85.14 &                       95.05 &                    77.71 &              2.05 &                             \textbf{47.37} &                             70.59 &                             85.45 &                             94.12 &                          78.95 &                    2.02 \\

    &            2 &              2 &                 2 &  230 &                       \textbf{47.39} &                       69.57 &                       80.00 &                       90.00 &                    73.48 &              2.14 &                             46.09 &                             70.43 &                             83.04 &                             92.61 &                          75.22 &                    2.08 \\

    &  2 &              2 &                 3 &           183 &              44.26 &                       63.39 &                       77.60 &                       92.35 &                    66.12 &              2.26 &                             45.90 &                             62.84 &                             79.78 &                             93.44 &                          71.04 &                    2.18 \\

    &  4 &              2 &                 0 &           230 &             50.87 &                       69.57 &                       82.61 &                       94.35 &                    73.48 &              2.05 &                             \textbf{50.87} &                             75.65 &                             86.52 &                             95.22 &                          74.78 &                    1.94 \\

& 4 &              2 &                 1 &             183 &          \textbf{47.54} &                       68.31 &                       83.06 &                       94.54 &                    71.04 &              2.11 &                             53.55 &                             72.68 &                             81.42 &                             92.90 &                          69.95 &                    2.06 \\

& 4 &              2 &                 2 &         102 &              42.16 &                       66.67 &                       83.33 &                       94.12 &                    71.57 &              2.19 &                             40.20 &                             59.80 &                             78.43 &                             93.14 &                          66.67 &                    2.35 \\

       & 2 &              4 &                 0 &    51 &                   41.18   &                       74.51 &                       84.31 &                       92.16 &                    78.43 &              2.11 &                             \textbf{56.86} &                             86.27 &                             90.20 &                             96.08 &                          84.31 &                    1.70     \\
    & & & & & & & & & & & & \\
     \multirow{7}{*}{GC} 
      & 2 &              2 &                 0 &    385 &                   70.13 &                       88.05 &                         - &                         - &                      - &              1.42 &                             70.13 &                             88.83 &                               - &                               - &                            - &                    1.41 \\

      & 2 &              2 &                 1 &      323 &                 56.97 &                       83.28 &                        -&                        -&                     -&              1.60 &                             51.39 &                             81.11 &                              -&                              -&                           -&                    1.67 \\

      & 2 &              2 &                 2 &          230 &             46.52 &                       76.09 &                        -&                        -&                     -&              1.77 &                             50.43 &                             81.74 &                              -&                              -&                           -&                    1.68 \\

      & 2 &              2 &                 3 &        183 &               48.09 &                       74.32 &                        -&                        -&                     -&              1.78 &                             44.81 &                             73.77 &                              -&                              -&                           -&                    1.81 \\

     & 4 &              2 &                 0 &        230 &                62.61 &                       86.52 &                        -&                        -&                     -&              1.51 &                             63.48 &                             85.65 &                              -&                              -&                           -&                    1.51 \\
     &        4 &              2 &                 1 &   183 &                    50.82 &                       82.51 &                         - &                         - &                      - &              1.67 &                             56.28 &                             84.15 &                               - &                               - &                            - &                    1.60 \\

      &4 &              2 &                 2 &    102 &                   45.10 &                       74.51 &                        -&                        -&                     -&              1.80 &                             39.22 &                             75.49 &                              -&                              -&                           -&                    1.85 \\

      &     2 &              4 &                 1 &      51   &              78.43 &                       90.20 &                        -&                        -&                     -&              1.31 &                             82.35 &                             90.20 &                              -&                              -&                           -&                    1.27 \\

    \Xhline{3\arrayrulewidth}
     \multicolumn{8}{l}{\textbf{MDC Test Corpus}} \\
    %\textbf{MDC Test Corpus} & & & & & & & \\
     %& & & & & & & \\
     \multirow{3}{*}{IEC} 
    &         2 &              2 &                 0 &    234 &                   52.99 &                       79.91 &                       90.60 &                       97.01 &                    85.47 &              1.79 &                             50.85 &                             74.79 &                             89.74 &                             96.15 &                          83.33 &                    1.88 \\
      &        2 &              2 &                 1 & 161 &                        66.46 &                       78.88 &                       91.93 &                       95.65 &                    86.34 &              1.67 &                             67.08 &                             82.61 &                             91.93 &                             95.03 &                          88.20 &                    1.63 \\
    &        2 &              2 &                 2 &         106 &              48.11 &                       72.64 &                       88.68 &                       95.28 &                    81.13 &              1.95 &                             47.17 &                             71.70 &                             85.85 &                             94.34 &                          79.25 &                    2.01 \\
     &        3 &              2 &                 0 &       161 &                  66.46 &                       78.88 &                       91.93 &                       95.65 &                    86.34 &              1.52 &                             67.08 &                             82.61 &                             91.93 &                             95.03 &                          88.20 &                    1.46 \\
     &        3 &              2 &                 1 &      106 &                 48.11 &                       72.64 &                       88.68 &                       95.28 &                    81.13 &              1.80 &                             47.17 &                             71.70 &                             85.85 &                             94.34 &                          79.25 &                    1.83 \\
      &        3 &              2 &                 2 &        75 &               \textbf{56.00} &                       81.33 &                       92.00 &                       96.00 &                    82.67 &              1.61 &                             56.00 &                             81.33 &                             92.00 &                             94.67 &                          88.00 &                    1.58 \\
    & & & & & & & & & & & & \\
    \multirow{3}{*}{IEC \& CU} 
    &          2 &              2 &                 0 &        234 &                 \textbf{65.81} &                       86.32 &                       93.59 &                       97.86 &                    93.16 &              1.56 &                             60.68 &                             82.48 &                             94.87 &                             98.72 &                          92.31 &                    1.63 \\
   &        2 &              2 &                 1 &         161 &                \textbf{67.08} &                       77.02 &                       90.06 &                       96.27 &                    84.47 &              1.70 &                             65.22 &                             80.75 &                             90.06 &                             95.03 &                          85.71 &   1.69 \\
    &        2 &              2 &                 2 &         106 &              \textbf{51.89} &                       69.81 &                       86.79 &                       96.23 &                    81.13 &              1.95 &                             50.00 &                             72.64 &                             88.68 &                             93.40 &                          78.30 &                    1.95  \\
    &        3 &              2 &                 0 &        161 &                 \textbf{68.32} &                       80.12 &                       93.17 &                       96.27 &                    85.71 &              1.49 &                             52.80 &                             75.78 &                             82.61 &                             95.65 &                          80.75 &                    1.79 \\
    &         3 &              2 &                 1 &    106 &                   \textbf{50.94} &                       68.87 &                       85.85 &                       95.28 &                    80.19 &              1.84 &                             42.45 &                             61.32 &                             77.36 &                             91.51 &                          81.13 &                    2.02 \\
    &        3 &              2 &                 2 &     75 &                  46.67 &                       66.67 &                       81.33 &                       94.67 &                    78.67 &              1.94 &                             28.00 &                             50.67 &                             73.33 &                             85.33 &                          58.67 &                    2.22 \\
    & & & & & & & & & & & & \\
     \multirow{3}{*}{GC}
     &         2 &              2 &                 0 &    234 &                     81.20 &                       95.73 &                        -&                        -&                     -&              1.23 &                             76.92 &                             95.30 &                              -&                              -&                           -&                    1.28 \\
     &        2 &              2 &                 1 &    161 &                   67.70 &                       88.20 &                        -&                        -&                     -&              1.44 &                             45.96 &                             79.50 &                              -&                              -&                           -&                    1.75 \\
    &        2 &              2 &                 2 &     106 &                  50.00 &                       84.91 &                        -&                        -&                     -&              1.65 &                             45.28 &                             66.98 &                              -&                              -&                           -&                    1.88 \\
     &        3 &              2 &                 0 &     161 &                    72.67 &                       90.06 &                        -&                        -&                     -&              1.37 &                             39.13 &                             69.57 &                              -&                              -&                           -&                    1.91 \\
     &        3 &              2 &                 1 &       106 &                46.23 &                       83.96 &                        -&                        -&                     -&              1.70 &                             48.11 &                             67.92 &                              -&                              -&                           -&                    1.84 \\
       &        3 &              2 &                 2 &    75 &                   45.33 &                       72.00 &                        -&                        -&                     -&              1.83 &                             24.00 &                             41.33 &                              -&                              -&                           -&                    2.35 \\
        %&        4 &              2 &                 0 &                       69.81 &                       89.62 &                        -&                        -&                     -&              1.41 &                             59.43 &                             86.79 &                              -&                              -&                           -&                    1.54 \\
        %&        4 &              2 &                 1 &                       52.00 &                       77.33 &                        -&                        -&                     -&              1.71 &                             34.67 &                             69.33 &                              -&                              -&                           -&                    1.96 \\
       % &        4 &              2 &                 2 &                       71.43 &                       82.86 &                        -&                        -&                     -&              1.46 &                             40.00 &                             71.43 &                              -&                              -&                           -&                    1.89 \\
    \Xhline{3\arrayrulewidth}
    \end{tabular}
    }
    \caption{Detailed Long-Term Planning Evaluation with $n=$ number of evaluation samples}
    \label{tab:LTPEval}
    %}
\end{table*}

\section{Ablation Study}\label{sec:ab}
As an ablation study, we compare two variations of a simple contrastive to our introduced curved contrastive objective. The first variation has the exact same setup as our approach with the same mixed learning objective of NLI, a dialogue window of $l=5$, the same hard negatives (including ones for the directional property) but without the "curved" similarity scores between \texttt{[BEFORE]} and \texttt{[AFTER]} tokens. In other words with simple labels of $0$ (not before and after each other within 5 turns) or $1$ (before and after the utterance with a distance between 1-5 turns). Since this does not take any distance into account we have a second ablation variant that takes only direct utterance pairs (so a window size of 2) with the corresponding two labels and otherwise the same setup. Like our embeddings, we train the two variations on BERT and RoBERTa architectures respectively. In contrast to our embeddings, we find that both ablation studies find their optimum for our three takes after already 1-2 epochs. In the following sections, we present the performance of the ablation studies to our approach, note that we refer to the ablation with a window size of $l=5$ as \texttt{ab5} and the one with $l=2$ as $\texttt{ab2}$.

\subsection{Ablation Study LTP}\label{sec:altp}
As shown in table \ref{tab:abLTPEvalShort} the ablation study with a dialogue window of $l=5$ shows stronger performance in ordering utterances than its counterpart of $l=2$.  Thanks to the solidified structure of the task-oriented corpus the ablation comes relatively close to the performance of our imaginary embeddings. For Greedy Curving (GC) in particular, it can detect the next goal out of 3 even slightly better than our embeddings without speaker tokens. However, when the solidified structure of dialogue disappears (on the chit-chat dataset DailyDialog) our models show much stronger performance than their ablation study.

\subsection{Ablation Study STP}\label{sec:astp}
While the ablation study with the dialogue window of $l=5$ shows solid performance in ordering utterances, it has severe trouble understanding the pathways between utterances as can be seen in \ref{tab:abSTPEvalShort}. Especially, on the MDC dataset for close members in their own group (observation window). Here we observe that the performances increase over longer distances which goes hand in hand with the better greedy curving performance. Overall, the ablation study with a dialogue window of $l=2$ shows through its learning objective a better understanding of its close neighbors as $l=5$. While once again the ablation studies do not get close to our embeddings on the DailyDialog corpus, on the MDC corpus it can outperform our embeddings on direct neighbors (distance $1$) while being significantly worse on longer distances. Since it only learned the properties between two speakers it has notable trouble mapping utterances from the same speaker as can be seen by even distances on the MDC corpus.

    \begin{table*}[!htbp]
\centering
 %\medium{
    \renewcommand{\arraystretch}{1.0}
    
    \centering
    
    \resizebox{\textwidth}{!}{
    
    \begin{tabular}{l | c c c c c c | c c c c c c}
    %\Xhline{3\arrayrulewidth}
    \multicolumn{5}{c}{}{\textbf{Ablation Study}}  &  & &   \multicolumn{5}{c}{}{\textbf{Imaginary Embedding with / w.o Speaker Token}}  \\

      & \multicolumn{3}{l}{}{\textbf{partially ordered}} & & \textbf{\thead{Reverse \\ order}} & & \multicolumn{3}{l}{}{\textbf{partially ordered}} &  & \textbf{\thead{Reverse \\ order}} \\
     
    %\Xhline{3\arrayrulewidth}
    \multirow{2}{*}{\textbf{Model}}& 
    \multirow{2}{*}{\textbf{\thead{Hits@1 \\ (in \%)}}} & 
    \multirow{2}{*}{\textbf{\thead{Hits@2 \\ (in \%)}}} & 
    \multirow{2}{*}{\textbf{\thead{Hits@3 \\ (in \%)}}} & 
    \multirow{2}{*}{\textbf{\thead{Hits@4 \\ (in \%)}}} &
    \multirow{2}{*}{\textbf{\thead{Hits@1 \\ (in \%)}}} &
    \multirow{2}{*}{\textbf{\thead{Average\\ Rank}}} &
    \multirow{2}{*}{\textbf{\thead{Hits@1 \\ (in \%)}}} & 
    \multirow{2}{*}{\textbf{\thead{Hits@2 \\ (in \%)}}} & 
    \multirow{2}{*}{\textbf{\thead{Hits@3 \\ (in \%)}}} & 
    \multirow{2}{*}{\textbf{\thead{Hits@4 \\ (in \%)}}} &
    \multirow{2}{*}{\textbf{\thead{Hits@1 \\ (in \%)}}} &
    \multirow{2}{*}{\textbf{\thead{Average\\ Rank}}} \\
    & & & & & & & & & & & & \\
     & & & & & & & & & & & & \\
    \Xhline{3\arrayrulewidth}
    %\Xhline{3\arrayrulewidth}
    \multicolumn{8}{l}{\textbf{DailyDialog Test Corpus }} \\
     %\textbf{DailyDialogue Test Corpus} & & & & & & & \\
         %& & & & & & & & & &\\
                  %& & & & & & & & & &\\
             IEC \texttt{ab5} / \texttt{ST}  &                      33.69 &                       56.30 &                       78.26 &                       90.00 &                   68.04 &               2.74    &                             \textbf{51.60} &                             72.22 &                             86.82 &                             94.94 &                          81.18 &                    \textbf{2.13}   \\
             IEC \texttt{ab2} /  \texttt{w.o} &                        25.69 &                       48.69 &                       67.43 &                       84.53 &                    58.31 &               3.15    & 49.99                             &  70.62                            &  85.26                            &        93.42                    & 79.17                          &      2.21               \\
     %\Xhline{3\arrayrulewidth}
     %\hline[dashed]
     & & & & & & & & & & & & \\applies to Greedy Curving where
   IEC \& CU  \texttt{ab5} / \texttt{ST} &                     33.48 &                       56.52 &                       79.13 &                       89.13 &                   69.13  &              2.73 &                                                   \textbf{51.07} &                             72.98 &                              86.9 &                             94.97 &                          79.87 &                    \textbf{2.13} \\
   IEC \& CU  \texttt{ab2} / \texttt{w.o} &26.71 &                       49.25 &                       69.38 &                       85.28 &                    60.03 &              3.09  &          50.69                  &      71.24                        &      85.09                       &    93.63                         &    78.54                     &     2.19               \\
    & & & & & & & & & & & & \\
    GC \texttt{ab5} / \texttt{ST} &                       50.00 &                       75.22 &                         - &                         - &                      - &               1.75 &                             57.32 &                             83.89 &                               - &                               - &                            - &                     \textbf{1.59} \\
     GC \texttt{ab2} / \texttt{w.o} &                       43.69 &                       72.68 &                         - &                         - &                      - &               1.84 &                          \textbf{57.87}    &                 82.47             &                               - &                               - &                            - &      1.6               \\
     
    \Xhline{3\arrayrulewidth}

         \multicolumn{8}{l}{\textbf{MDC Test Cor pus}} \\
    %\textbf{MDC Test Corpus} & & & & & & & \\
     %& & & & & & & \\
     IEC        \texttt{ab5} / \texttt{ST}     &           54.62 &                      74.15  &                      89.42 &                       96.5 &                   84.71  &              2.01 &                             56.83 &                             77.50 &                             90.19 &                             95.44 &                          84.52 &                    1.96 \\
      IEC \texttt{ab2} / \texttt{w.o} &                       41.52 &                       65.19 &                       84.50 &                       94.17 &                    75.61 &             2.39  &                             \textbf{58.72} &                             77.50 &                             90.19 &                             95.44 &                          84.52 &         \textbf{1.92}            \\
     & & & & & & & & & \\
    IEC \& CU  \texttt{ab5} / \texttt{ST} &                       54.77 &                       75.02 &                       89.91 &                       96.97 &                    85.45 &              1.98 &                             58.63 &                             78.62 &                             91.20 &                             95.72 &                          85.44 &                    1.90 \\
    IEC \& CU  \texttt{ab2} / \texttt{w.o} &                       40.83 &                       64.24 &                       84.38 &                       94.05 &                    75.86 &              2.41 &                             \textbf{61.59} &                             77.72 &                             90.15 &                             96.79 &                          86.25 &                    \textbf{1.87} \\
     & & & & & & & & & \\
     GC  \texttt{ab5} / \texttt{ST}&                       \textbf{66.63} &                       89.86 &                         - &                         - &                      - &              \textbf{1.43} &                             56.05 &                             80.59 &                               - &                               - &                            - &                    1.64 \\
     GC  \texttt{ab2} / \texttt{w.o} &                       48.77 &                       72.68 &                         - &                         - &                      - &              1.72 &                             ´66.30 &                             89.61 &                               - &                               - &                            - &                    1.44 \\
    \Xhline{3\arrayrulewidth}

    \end{tabular}
    }
    \caption{Aggregated Long-Term Planning Ablation vs Imaginary Embeddings Study on 3 goals with ((2, 2, 2), (2, 2, 0) and (2, 2, 1)) with (history length, goal distances, first goal \textbf{in} distance). Models include Imaginary Embedding Chain (IEC),  Imaginary Embedding Chain + Curving (IEC \& CU), and Greedy Curving (GC). (\texttt{ab2} ablation with $l=2$), (\texttt{ab5} ablation with $l=5$),  (\texttt{w.o} without Speaker Token), (\texttt{ST} with Speaker Token)  }
    \label{tab:abLTPEvalShort}
    %}
\end{table*}
\begin{table*}[!htbp]
\centering

 %\medium{
    \renewcommand{\arraystretch}{1.2}
    
    \centering
    %\label{tab:STPEvalShort}
    \resizebox{\textwidth}{!}{
    
    \begin{tabular}{l | c c c c c | c c c c c}
    \Xhline{3\arrayrulewidth}
     & & & \multicolumn{6}{c}{}{\textbf{\thead{Human Utterance Ranking vs 100 utterances sampled \\ from DialoGPT Large / GODEL Large (p=0.8, t=0.8)}}} \\

      & & \multicolumn{3}{c}{}{\textbf{\thead{Ablation Study}}} & & & \multicolumn{3}{c}{}{\textbf{\thead{Imaginary Embedding \\ with (\texttt{ST}) / \texttt{w.o} Speaker Token}}}  \\
    %\Xhline{3\arrayrulewidth}
    \multirow{2}{*}{\textbf{Goal in Distance}}& 
    \multirow{2}{*}{\textbf{\thead{Hits@5 \\ (in \%)}}} & 
    \multirow{2}{*}{\textbf{\thead{Hits@10 \\ (in \%)}}} & 
    \multirow{2}{*}{\textbf{\thead{Hits@25 \\ (in \%)}}} & 
    \multirow{2}{*}{\textbf{\thead{Hits@50 \\ (in \%)}}} &
    \multirow{2}{*}{\textbf{\thead{Average \\ Rank}}} &
    \multirow{2}{*}{\textbf{\thead{Hits@5 \\ (in \%)}}} & 
    \multirow{2}{*}{\textbf{\thead{Hits@10 \\ (in \%)}}} & 
    \multirow{2}{*}{\textbf{\thead{Hits@25 \\ (in \%)}}} & 
    \multirow{2}{*}{\textbf{\thead{Hits@50 \\ (in \%)}}} &
    \multirow{2}{*}{\textbf{\thead{Average \\ Rank}}} \\
    & & & & & & & & & & \\
    \Xhline{3\arrayrulewidth}
    %\Xhline{3\arrayrulewidth}
    %\multicolumn{8}{l}{\textbf{DailyDialogue Test Corpus even goal distance $g_d$ (saying goal by yourself)}} \\
     \textbf{DailyDialog Test Corpus} & & & & &  \\
        
         Guidance distance $1$ (\texttt{ab5} /  \texttt{ST}) &  21.87      &    29.96     &     45.27   &     63.63     &           39.34   &          \textbf{72.03} &           79.53 &           88.37 &           94.84 &                  \textbf{8.82}   \\
        Guidance distance  $1$ (\texttt{ab2} /  \texttt{w.o}) &      45.57 &       52.83 &       68.03 &        82.27 &             22.27 &     38.06       &      46.26      &       59.86     &    77.00        &     26.70                \\
                  & & & & & & & & & &  \\
        Guidance distance $2$ (\texttt{ab5} /  \texttt{ST}) &    19.70    &  26.47       &    42.57    &     61.94     &          40.71    &          27.28 &           36.07 &           55.50 &           71.89 &                 31.83  \\
        Guidance distance  $2$ (\texttt{ab2} /  \texttt{w.o}) &      20.22 &       25.76 &       41.97 &        59.40 &            43.89 &      \textbf{32.06}    &    38.31        &   51.20         &    70.46        &         32.94            \\
                  & & & & & & & & & &  \\

        Guidance distance $3$ (\texttt{ab5} /  \texttt{ST}) &    14.69    &    22.88     &   38.87     &   62.31      &            41.53 &         \textbf{50.68} &           61.17 &           75.46 &           85.38 &                 19.01   \\
        Guidance distance  $3$ (\texttt{ab2} /  \texttt{w.o}) &     20.21 &       27.55 &       43.135 &        63.35 &            39.96 &        21.18     &     28.62       &    45.42        &      66.47      &         36.47            \\
                  & & & & & & & & & &  \\

        Guidance distance $4$ (\texttt{ab5} /  \texttt{ST}) &     15.25   &    22.54     &     35.94   &      57.16    &          45.60    &         \textbf{28.28} &           36.37 &                    52.05 &        70.83 &           33.28   \\
        Guidance distance  $4$ (\texttt{ab2} /  \texttt{w.o}) &     19.67 &       25.15 &       38.64 &        57.18 &            44.40  &      26.67     &    33.22        &  50.06          &       65.35     &      36.23               \\
                  & & & & & & & & & &  \\

            \Xhline{3\arrayrulewidth}

     \textbf{MDC Test Corpus} & & & & &  \\
         %& & & & & & & & & &  \\
        Guidance distance $1$ (\texttt{ab5} /  \texttt{ST}) &    6.99   &  11.00    &    23.62 &    43.88    & 54.94            &         61.48 &           68.97 &          79.32 &           88.59 &                 14.94  \\
         Guidance distance $1$ (\texttt{ab2} /  \texttt{w.o}) &      \textbf{66.12} &      74.48 &      88.48 &        95.70 &           \textbf{9.09} &         29.59 &           36.12 &          49.31 &           68.75 &                 33.83  \\
& & & & & & & & & &  \\

    Guidance distance $2$ (\texttt{ab5} /  \texttt{ST}) &   9.15    &   15.012   &  31.72     &   53.33     &     47.52       &       \textbf{33.55} &     45.27       &     64.85       &   80.40         &   \textbf{24.86} \\
         Guidance distance $2$ (\texttt{ab2} /  \texttt{w.o}) & 4.1 &       6.29 &       14.24 &        32.13 &            63.97 &         20.84 &    28.22       &           45.27 &           67.90 &                 36.13   \\
& & & & & & & & & &  \\

    Guidance distance $3$ (\texttt{ab5} /  \texttt{ST}) &   8.40     &   12.92   &   28.735   &   51.41     &   49.50         &           \textbf{65.03} &           72.74 &          82.96 &           89.96 &                  \textbf{12.84}   \\
         Guidance distance $3$ (\texttt{ab2} /  \texttt{w.o}) & 25.01 &       32.58 &       51.56 &         68.88 &            33.97 &          20.18 &    27.13        &        43.66    &    65.50        &    38.44               \\
& & & & & & & & & &  \\

Guidance distance $4$ (\texttt{ab5} /  \texttt{ST}) &    13.50   &  19.73    &  28.75    &      56.39  &     43.98       &          \textbf{44.82} &           56.53 &          73.73 &           85.80 &                  \textbf{19.31}  \\
         Guidance distance $4$ (\texttt{ab2} /  \texttt{w.o})&      2.95 &       3.76 &       9.34 &        21.3 &            73.7 &          20.73 &     30.42        &  50.80                    &    73.80      &      33.59        \\

    %& 10 & 4 &  17.65 &   19.61 &   35.29 &   47.06 &  50.41 \\
    %& & & & & & & \\
    %\Xhline{3\arrayrulewidth}
     %& & & & & & & & & & \\

    %& & & & & & & \\

    \Xhline{3\arrayrulewidth}
    \end{tabular}
    }
    \caption{Aggregated short-term planning evaluation vs ablation study for different distances to goal. (\texttt{ab2} ablation with $l=2$), (\texttt{ab5} ablation with $l=5$),  (\texttt{w.o} without Speaker Token), (\texttt{ST} with Speaker Token)}
    \label{tab:abSTPEvalShort}
    %}
\end{table*}

%\newpage

\subsection{Ablation Study Next Utterance Selection} \label{sec:abNext}
\begin{figure}[hbt!]
    \centering
    \includegraphics[width=0.49\textwidth]{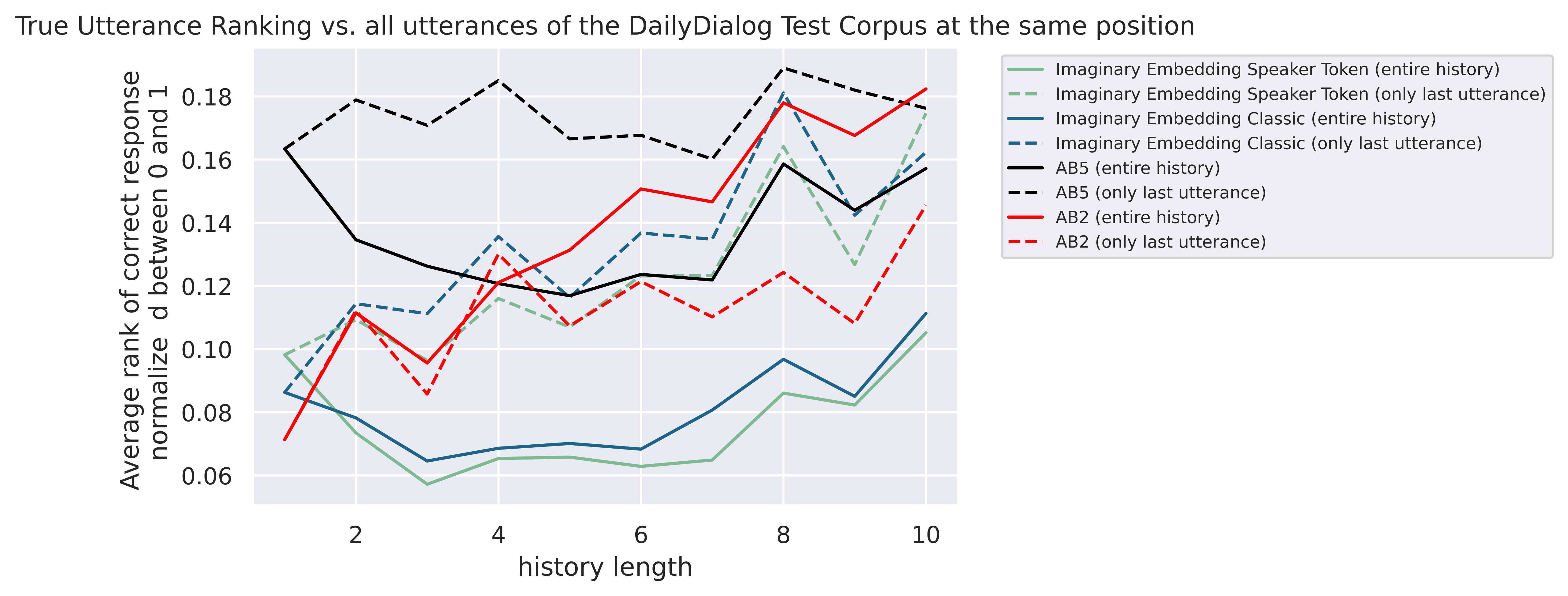}
    \caption{Normalized average rank of next utterance selection based on dialogue history on DailyDialog. Demonstrated are different Curving variants (only the last utterance or the entire history), classic as well as Speaker Token-based embeddings. As baselines, we utilize the two ablation study variants with the two variations' entire context or only the last utterance.}
    \label{fig:abrobust}
\end{figure}

\begin{figure}[hbt!]
    \centering
    \includegraphics[width=0.49\textwidth]{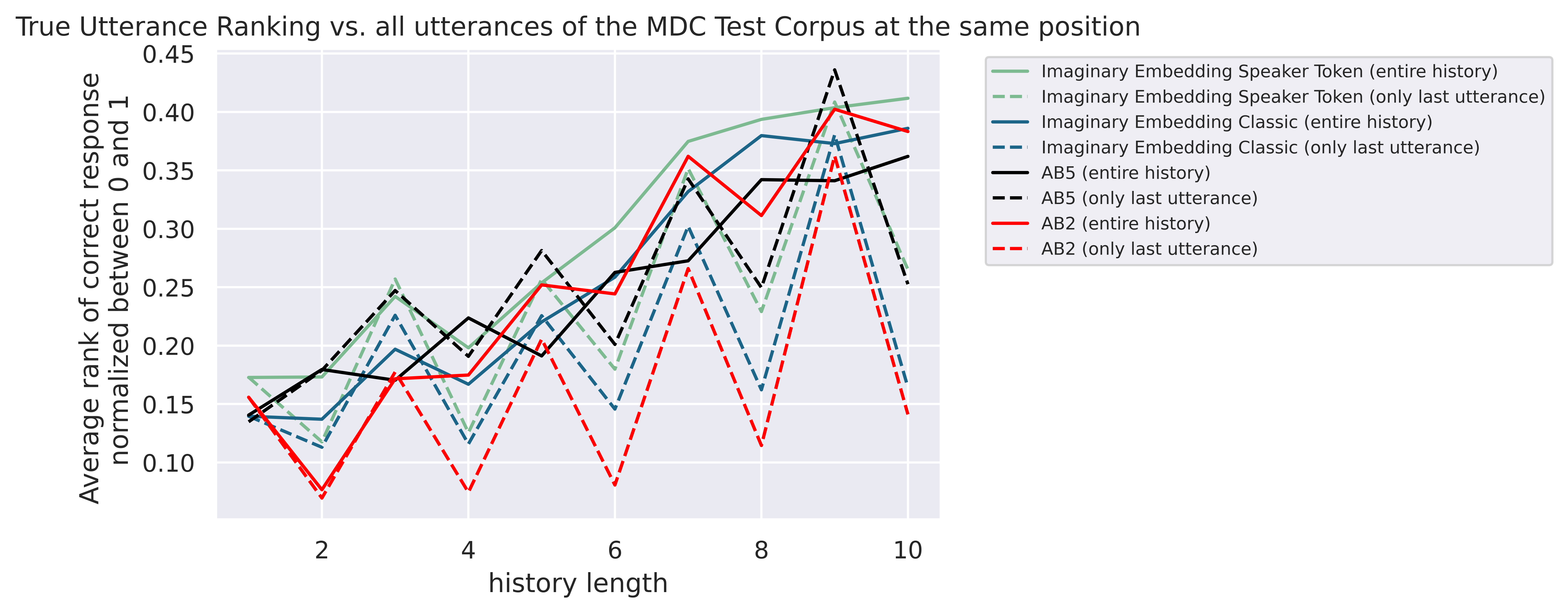}
    \caption{Normalized average rank of next utterance selection based on dialogue history on MDC. Demonstrated are different Curving variants (only the last utterance or the entire history), classic as well as Speaker Token-based embeddings. As baselines, we utilize the two ablation study variants with the two variations' entire context or only the last utterance.}
    \label{fig:abrobustMDC}
\end{figure}

We compare both ablation studies to our embeddings in figure \ref{fig:abrobust} on DailyDialog on the same variation as Imaginary Embeddings, either the entire context or only the last utterance. 
Both ablation studies perform best on the variation closest to their training target, in other words, ab5 on the entire context and ab2 only on the last utterance. With the Greedy Curving evaluation (table \ref{tab:abLTPEvalShort}), one could suggest a stronger performance to ab5 rather than ab2.
However, we find the exact opposite in the next utterance selection task as we consider candidate utterances in width rather than in-depth. Compared to the other baselines, the strongest ablation study is still 1.5\% worse than the pre-trained DialogRPT, 3.69\% worse than ConveRT, and 4.3\% worse than our best imaginary embeddings. On MDC (figure \ref{fig:abrobustMDC}), we observe, as we described in §\ref{sec:NextE}, that considering only the last utterance shows the strongest results. Expectedly, the training objective to only match direct pairs of ablation $l=2$ comes in handy, outperforming all other approaches.

\begin{thebibliography}{4}
\expandafter\ifx\csname natexlab\endcsname\relax\def\natexlab#1{#1}\fi

\bibitem[{Ando and Zhang(2005)}]{Ando2005}
Rie~Kubota Ando and Tong Zhang. 2005.
\newblock A framework for learning predictive structures from multiple tasks
  and unlabeled data.
\newblock \emph{Journal of Machine Learning Research}, 6:1817--1853.

\bibitem[{Andrew and Gao(2007)}]{andrew2007scalable}
Galen Andrew and Jianfeng Gao. 2007.
\newblock Scalable training of {L1}-regularized log-linear models.
\newblock In \emph{Proceedings of the 24th International Conference on Machine
  Learning}, pages 33--40.

\bibitem[{Gusfield(1997)}]{Gusfield:97}
Dan Gusfield. 1997.
\newblock \emph{Algorithms on Strings, Trees and Sequences}.
\newblock Cambridge University Press, Cambridge, UK.

\bibitem[{Rasooli and Tetreault(2015)}]{rasooli-tetrault-2015}
Mohammad~Sadegh Rasooli and Joel~R. Tetreault. 2015.
\newblock \href {http://arxiv.org/abs/1503.06733} {Yara parser: {A} fast and
  accurate dependency parser}.
\newblock \emph{Computing Research Repository}, arXiv:1503.06733.
\newblock Version 2.

\end{thebibliography}


\begin{thebibliography}{35}
\expandafter\ifx\csname natexlab\endcsname\relax\def\natexlab#1{#1}\fi

\bibitem[{Alderson-Day and Fernyhough(2015)}]{AldersonDay2015InnerSD}
Ben Alderson-Day and Charles Fernyhough. 2015.
\newblock Inner speech: Development, cognitive functions, phenomenology, and
  neurobiology.
\newblock \emph{Psychological Bulletin}, 141:931 -- 965.

\bibitem[{Bowman et~al.(2015)Bowman, Angeli, Potts, and
  Manning}]{https://doi.org/10.48550/arxiv.1508.05326}
Samuel~R. Bowman, Gabor Angeli, Christopher Potts, and Christopher~D. Manning.
  2015.
\newblock \href {https://doi.org/10.48550/ARXIV.1508.05326} {A large annotated
  corpus for learning natural language inference}.

\bibitem[{Chen et~al.(2022)Chen, Feng, and
  Zhao}]{https://doi.org/10.48550/arxiv.2204.03286}
Zhibin Chen, Yansong Feng, and Dongyan Zhao. 2022.
\newblock \href {https://doi.org/10.48550/ARXIV.2204.03286} {Entailment graph
  learning with textual entailment and soft transitivity}.

\bibitem[{Devlin et~al.(2018)Devlin, Chang, Lee, and
  Toutanova}]{https://doi.org/10.48550/arxiv.1810.04805}
Jacob Devlin, Ming-Wei Chang, Kenton Lee, and Kristina Toutanova. 2018.
\newblock \href {https://doi.org/10.48550/ARXIV.1810.04805} {Bert: Pre-training
  of deep bidirectional transformers for language understanding}.

\bibitem[{Dunbar et~al.(1997)Dunbar, Marriott, and Duncan}]{articleRelation}
Robin Dunbar, Anna Marriott, and Neill Duncan. 1997.
\newblock \href {https://doi.org/10.1007/BF02912493} {Human conversational
  behavior}.
\newblock \emph{Human nature (Hawthorne, N.Y.)}, 8:231--246.

\bibitem[{Dziri et~al.(2019)Dziri, Kamalloo, Mathewson, and
  Za{\"{\i}}ane}]{EvaluatingCoherence}
Nouha Dziri, Ehsan Kamalloo, Kory~W. Mathewson, and Osmar~R. Za{\"{\i}}ane.
  2019.
\newblock \href {http://arxiv.org/abs/1904.03371} {Evaluating coherence in
  dialogue systems using entailment}.
\newblock \emph{CoRR}, abs/1904.03371.

\bibitem[{Einstein(1921)}]{Einstein1921-EINRTS-2}
Albert Einstein. 1921.
\newblock \emph{Relativity: The Special and General Theory}.
\newblock Routledge.

\bibitem[{Gao et~al.(2020)Gao, Zhang, Galley, Brockett, and
  Dolan}]{DBLP:journals/corr/abs-2009-06978}
Xiang Gao, Yizhe Zhang, Michel Galley, Chris Brockett, and Bill Dolan. 2020.
\newblock \href {http://arxiv.org/abs/2009.06978} {Dialogue response ranking
  training with large-scale human feedback data}.
\newblock \emph{CoRR}, abs/2009.06978.

\bibitem[{Grandchamp et~al.(2019)Grandchamp, Rapin, Perrone-Bertolotti, Pichat,
  Haldin, Cousin, Lachaux, Dohen, Perrier, Garnier, Baciu, and
  Loevenbruck}]{grandchamp:hal-02290943}
Romain Grandchamp, Lucile Rapin, Marcela Perrone-Bertolotti, C{\'e}dric Pichat,
  C{\'e}lise Haldin, Emilie Cousin, Jean-Philippe Lachaux, Marion Dohen, Pascal
  Perrier, Ma{\"e}va Garnier, Monica Baciu, and H{\'e}l{\`e}ne Loevenbruck.
  2019.
\newblock \href {https://doi.org/10.3389/fpsyg.2019.02019} {{The ConDialInt
  Model: Condensation, Dialogality, and Intentionality Dimensions of Inner
  Speech Within a Hierarchical Predictive Control Framework}}.
\newblock \emph{{Frontiers in Psychology}}, 10:2019.

\bibitem[{Henderson et~al.(2020)Henderson, Casanueva, Mrk{\v{s}}i{\'c}, Su,
  Wen, and Vuli{\'c}}]{henderson-etal-2020-convert}
Matthew Henderson, I{\~n}igo Casanueva, Nikola Mrk{\v{s}}i{\'c}, Pei-Hao Su,
  Tsung-Hsien Wen, and Ivan Vuli{\'c}. 2020.
\newblock \href {https://doi.org/10.18653/v1/2020.findings-emnlp.196}
  {{C}onve{RT}: Efficient and accurate conversational representations from
  transformers}.
\newblock In \emph{Findings of the Association for Computational Linguistics:
  EMNLP 2020}, pages 2161--2174, Online. Association for Computational
  Linguistics.

\bibitem[{Johnson et~al.(2017)Johnson, Douze, and
  J{\'{e}}gou}]{DBLP:journals/corr/JohnsonDJ17}
Jeff Johnson, Matthijs Douze, and Herv{\'{e}} J{\'{e}}gou. 2017.
\newblock \href {http://arxiv.org/abs/1702.08734} {Billion-scale similarity
  search with gpus}.
\newblock \emph{CoRR}, abs/1702.08734.

\bibitem[{Kotlerman et~al.(2015)Kotlerman, Dagan, Magnini, and
  Bentivogli}]{Kotlerman2015TextualEG}
Lili Kotlerman, Ido Dagan, Bernardo Magnini, and Luisa Bentivogli. 2015.
\newblock Textual entailment graphs.
\newblock \emph{Natural Language Engineering}, 21:699 -- 724.

\bibitem[{Li et~al.(2018)Li, Wang, Sun, Panda, Liu, and
  Gao}]{https://doi.org/10.48550/arxiv.1807.11125}
Xiujun Li, Yu~Wang, Siqi Sun, Sarah Panda, Jingjing Liu, and Jianfeng Gao.
  2018.
\newblock \href {https://doi.org/10.48550/ARXIV.1807.11125} {Microsoft dialogue
  challenge: Building end-to-end task-completion dialogue systems}.

\bibitem[{Li et~al.(2017)Li, Su, Shen, Li, Cao, and
  Niu}]{li-etal-2017-dailydialog}
Yanran Li, Hui Su, Xiaoyu Shen, Wenjie Li, Ziqiang Cao, and Shuzi Niu. 2017.
\newblock \href {https://aclanthology.org/I17-1099} {{D}aily{D}ialog: A
  manually labelled multi-turn dialogue dataset}.
\newblock In \emph{Proceedings of the Eighth International Joint Conference on
  Natural Language Processing (Volume 1: Long Papers)}, pages 986--995, Taipei,
  Taiwan. Asian Federation of Natural Language Processing.

\bibitem[{Li et~al.(2021)Li, Kiseleva, and
  de~Rijke}]{DBLP:journals/corr/abs-2105-00079}
Ziming Li, Julia Kiseleva, and Maarten de~Rijke. 2021.
\newblock \href {http://arxiv.org/abs/2105.00079} {Improving response quality
  with backward reasoning in open-domain dialogue systems}.
\newblock \emph{CoRR}, abs/2105.00079.

\bibitem[{Liu et~al.(2021)Liu, Wang, Liu, Sun, Huang, and
  Si}]{https://doi.org/10.48550/arxiv.2109.12599}
Che Liu, Rui Wang, Jinghua Liu, Jian Sun, Fei Huang, and Luo Si. 2021.
\newblock \href {https://doi.org/10.48550/ARXIV.2109.12599} {Dialoguecse:
  Dialogue-based contrastive learning of sentence embeddings}.

\bibitem[{Liu et~al.(2019)Liu, Ott, Goyal, Du, Joshi, Chen, Levy, Lewis,
  Zettlemoyer, and Stoyanov}]{https://doi.org/10.48550/arxiv.1907.11692}
Yinhan Liu, Myle Ott, Naman Goyal, Jingfei Du, Mandar Joshi, Danqi Chen, Omer
  Levy, Mike Lewis, Luke Zettlemoyer, and Veselin Stoyanov. 2019.
\newblock \href {https://doi.org/10.48550/ARXIV.1907.11692} {Roberta: A
  robustly optimized bert pretraining approach}.

\bibitem[{Mehrabi et~al.(2019)Mehrabi, Morstatter, Saxena, Lerman, and
  Galstyan}]{DBLP:journals/corr/abs-1908-09635}
Ninareh Mehrabi, Fred Morstatter, Nripsuta Saxena, Kristina Lerman, and Aram
  Galstyan. 2019.
\newblock \href {http://arxiv.org/abs/1908.09635} {A survey on bias and
  fairness in machine learning}.
\newblock \emph{CoRR}, abs/1908.09635.

\bibitem[{Meister et~al.(2022)Meister, Pimentel, Wiher, and
  Cotterell}]{DBLP:journals/corr/abs-2202-00666}
Clara Meister, Tiago Pimentel, Gian Wiher, and Ryan Cotterell. 2022.
\newblock \href {http://arxiv.org/abs/2202.00666} {Typical decoding for natural
  language generation}.
\newblock \emph{CoRR}, abs/2202.00666.

\bibitem[{Myllyniemi(1986)}]{MYLLYNIEMI1986147}
Rauni Myllyniemi. 1986.
\newblock \href {https://doi.org/https://doi.org/10.1016/0271-5309(86)90019-4}
  {Conversation as a system of social interaction}.
\newblock \emph{Language \& Communication}, 6(3):147--169.

\bibitem[{Obamuyide and Vlachos(2018)}]{obamuyide-vlachos-2018-zero}
Abiola Obamuyide and Andreas Vlachos. 2018.
\newblock \href {https://doi.org/10.18653/v1/W18-5511} {Zero-shot relation
  classification as textual entailment}.
\newblock In \emph{Proceedings of the First Workshop on Fact Extraction and
  {VER}ification ({FEVER})}, pages 72--78, Brussels, Belgium. Association for
  Computational Linguistics.

\bibitem[{Patterson et~al.(2021)Patterson, Gonzalez, Le, Liang, Munguia,
  Rothchild, So, Texier, and Dean}]{DBLP:journals/corr/abs-2104-10350}
David~A. Patterson, Joseph Gonzalez, Quoc~V. Le, Chen Liang, Lluis{-}Miquel
  Munguia, Daniel Rothchild, David~R. So, Maud Texier, and Jeff Dean. 2021.
\newblock \href {http://arxiv.org/abs/2104.10350} {Carbon emissions and large
  neural network training}.
\newblock \emph{CoRR}, abs/2104.10350.

\bibitem[{Peng et~al.(2022)Peng, Galley, He, Brockett, Liden, Nouri, Yu, Dolan,
  and Gao}]{peng2022godel}
Baolin Peng, Michel Galley, Pengcheng He, Chris Brockett, Lars Liden, Elnaz
  Nouri, Zhou Yu, Bill Dolan, and Jianfeng Gao. 2022.
\newblock \href
  {https://www.microsoft.com/en-us/research/publication/godel-large-scale-pre-training-for-goal-directed-dialog/}
  {Godel: Large-scale pre-training for goal-directed dialog}.
\newblock arXiv.

\bibitem[{Ramakrishnan et~al.(2022)Ramakrishnan, Narangodage, Schilman,
  Weinberger, and McDonald}]{https://doi.org/10.48550/arxiv.2205.07352}
Ramya Ramakrishnan, Hashan~Buddhika Narangodage, Mauro Schilman, Kilian~Q.
  Weinberger, and Ryan McDonald. 2022.
\newblock \href {https://doi.org/10.48550/ARXIV.2205.07352} {Long-term control
  for dialogue generation: Methods and evaluation}.

\bibitem[{Reimers and Gurevych(2019)}]{DBLP:journals/corr/abs-1908-10084}
Nils Reimers and Iryna Gurevych. 2019.
\newblock \href {http://arxiv.org/abs/1908.10084} {Sentence-bert: Sentence
  embeddings using siamese bert-networks}.
\newblock \emph{CoRR}, abs/1908.10084.

\bibitem[{Robertson and Zaragoza(2009)}]{10.1561/1500000019}
Stephen Robertson and Hugo Zaragoza. 2009.
\newblock \href {https://doi.org/10.1561/1500000019} {The probabilistic
  relevance framework: Bm25 and beyond}.
\newblock \emph{Found. Trends Inf. Retr.}, 3(4):333–389.

\bibitem[{Schramowski et~al.(2022)Schramowski, Turan, Andersen, Rothkopf, and
  Kersting}]{Schramowski2022}
Patrick Schramowski, Cigdem Turan, Nico Andersen, Constantin~A. Rothkopf, and
  Kristian Kersting. 2022.
\newblock \href {https://doi.org/10.1038/s42256-022-00458-8} {Large pre-trained
  language models contain human-like biases of what is right and wrong to do}.
\newblock \emph{Nature Machine Intelligence}, 4(3):258--268.

\bibitem[{Strubell et~al.(2019)Strubell, Ganesh, and
  McCallum}]{DBLP:journals/corr/abs-1906-02243}
Emma Strubell, Ananya Ganesh, and Andrew McCallum. 2019.
\newblock \href {http://arxiv.org/abs/1906.02243} {Energy and policy
  considerations for deep learning in {NLP}}.
\newblock \emph{CoRR}, abs/1906.02243.

\bibitem[{Teixeira and Dragoni(2022)}]{articlePlanning}
Milene Teixeira and Mauro Dragoni. 2022.
\newblock \href {https://doi.org/10.1007/s12559-022-09996-0} {A review of
  plan-based approaches for dialogue management}.
\newblock \emph{Cognitive Computation}, 14.

\bibitem[{Weidinger et~al.(2021)Weidinger, Mellor, Rauh, Griffin, Uesato,
  Huang, Cheng, Glaese, Balle, Kasirzadeh, Kenton, Brown, Hawkins, Stepleton,
  Biles, Birhane, Haas, Rimell, Hendricks, Isaac, Legassick, Irving, and
  Gabriel}]{DBLP:journals/corr/abs-2112-04359}
Laura Weidinger, John Mellor, Maribeth Rauh, Conor Griffin, Jonathan Uesato,
  Po{-}Sen Huang, Myra Cheng, Mia Glaese, Borja Balle, Atoosa Kasirzadeh, Zac
  Kenton, Sasha Brown, Will Hawkins, Tom Stepleton, Courtney Biles, Abeba
  Birhane, Julia Haas, Laura Rimell, Lisa~Anne Hendricks, William~S. Isaac,
  Sean Legassick, Geoffrey Irving, and Iason Gabriel. 2021.
\newblock \href {http://arxiv.org/abs/2112.04359} {Ethical and social risks of
  harm from language models}.
\newblock \emph{CoRR}, abs/2112.04359.

\bibitem[{Williams et~al.(2017)Williams, Nangia, and
  Bowman}]{https://doi.org/10.48550/arxiv.1704.05426}
Adina Williams, Nikita Nangia, and Samuel~R. Bowman. 2017.
\newblock \href {https://doi.org/10.48550/ARXIV.1704.05426} {A broad-coverage
  challenge corpus for sentence understanding through inference}.

\bibitem[{Yin et~al.(2019)Yin, Hay, and
  Roth}]{DBLP:journals/corr/abs-1909-00161}
Wenpeng Yin, Jamaal Hay, and Dan Roth. 2019.
\newblock \href {http://arxiv.org/abs/1909.00161} {Benchmarking zero-shot text
  classification: Datasets, evaluation and entailment approach}.
\newblock \emph{CoRR}, abs/1909.00161.

\bibitem[{You et~al.(2020)You, Chen, Sui, Chen, Wang, and
  Shen}]{NEURIPS2020_3fe23034}
Yuning You, Tianlong Chen, Yongduo Sui, Ting Chen, Zhangyang Wang, and Yang
  Shen. 2020.
\newblock \href
  {https://proceedings.neurips.cc/paper/2020/file/3fe230348e9a12c13120749e3f9fa4cd-Paper.pdf}
  {Graph contrastive learning with augmentations}.
\newblock In \emph{Advances in Neural Information Processing Systems},
  volume~33, pages 5812--5823. Curran Associates, Inc.

\bibitem[{Zeng et~al.(2021)Zeng, Wu, Hu, Xu, Yuan, Huang, Zhuang, Hu, and
  Shi}]{DBLP:journals/corr/abs-2106-09157}
Dewen Zeng, Yawen Wu, Xinrong Hu, Xiaowei Xu, Haiyun Yuan, Meiping Huang, Jian
  Zhuang, Jingtong Hu, and Yiyu Shi. 2021.
\newblock \href {http://arxiv.org/abs/2106.09157} {Positional contrastive
  learning for volumetric medical image segmentation}.
\newblock \emph{CoRR}, abs/2106.09157.

\bibitem[{Zhang et~al.(2019)Zhang, Sun, Galley, Chen, Brockett, Gao, Gao, Liu,
  and Dolan}]{DialogGPT}
Yizhe Zhang, Siqi Sun, Michel Galley, Yen-Chun Chen, Chris Brockett, Xiang Gao,
  Jianfeng Gao, Jingjing Liu, and Bill Dolan. 2019.
\newblock \href {https://doi.org/10.48550/ARXIV.1911.00536} {Dialogpt:
  Large-scale generative pre-training for conversational response generation}.

\end{thebibliography}
\end{document}